\documentclass[10pt,twocolumn,letterpaper]{article}

\usepackage[pagenumbers]{cvpr}              
\usepackage{graphicx}
\usepackage{amsmath}
\usepackage{amssymb}
\usepackage{booktabs}
\usepackage{multirow}
\usepackage{bbm}
\usepackage{adjustbox}

\usepackage{amsmath,amsfonts,bm}

\def\eqref#1{equation~\ref{#1}}

\def\1{\bm{1}}

\DeclareMathAlphabet{\mathsfit}{\encodingdefault}{\sfdefault}{m}{sl}
\SetMathAlphabet{\mathsfit}{bold}{\encodingdefault}{\sfdefault}{bx}{n}

\usepackage[usenames,dvipsnames]{xcolor}
\usepackage{xcolor}
\usepackage{pifont}

\usepackage[pagebackref,breaklinks,colorlinks]{hyperref}

\usepackage[capitalize]{cleveref}
\crefname{section}{Sec.}{Secs.}
\Crefname{section}{Section}{Sections}
\Crefname{table}{Table}{Tables}
\crefname{table}{Tab.}{Tabs.}

\def\cvprPaperID{11521} \def\confName{CVPR}
\def\confYear{2022}

\usepackage{overpic}
\usepackage{enumitem} 
\usepackage{overpic} 
\usepackage{color}

\definecolor{turquoise}{cmyk}{0.65,0,0.1,0.3}
\definecolor{purple}{rgb}{0.65,0,0.65}
\definecolor{dark_green}{rgb}{0, 0.5, 0}
\definecolor{orange}{rgb}{0.8, 0.6, 0.2}
\definecolor{red}{rgb}{0.8, 0.2, 0.2}
\definecolor{darkred}{rgb}{0.6, 0.1, 0.05}
\definecolor{blueish}{rgb}{0.0, 0.3, .6}
\definecolor{light_gray}{rgb}{0.7, 0.7, .7}
\definecolor{pink}{rgb}{1, 0, 1}
\definecolor{greyblue}{rgb}{0.25, 0.25, 1}

\usepackage{blindtext}

\renewcommand{\paragraph}[1]{\vspace{1em}\noindent\textbf{#1}.}
\begin{document}

\title{Investigating Top-$k$ White-Box and Transferable Black-box Attack}

\author{
      Chaoning Zhang, Philipp Benz, Adil Karjauv, Jae Won Cho, Kang Zhang,
      In So Kweon \\
     Korea Advanced Institute of Science and Technology (KAIST) \\
     chaoningzhang1990@gmail.com
}

\maketitle

\begin{abstract}
Existing works have identified the limitation of top-$1$ attack success rate (ASR) as a metric to evaluate the attack strength but exclusively investigated it in the white-box setting, while our work extends it to a more practical black-box setting: transferable attack. It is widely reported that stronger I-FGSM transfers worse than simple FGSM, leading to a popular belief that transferability is at odds with the white-box attack strength. Our work challenges this belief with empirical finding that stronger attack actually transfers better for the general top-$k$ ASR indicated by the interest class rank (ICR) after attack. For increasing the attack strength, with an intuitive interpretation of the logit gradient from the geometric perspective, we identify that the weakness of the commonly used losses lie in prioritizing the speed to fool the network instead of maximizing its strength. To this end, we propose a new normalized CE loss that guides the logit to be updated in the direction of implicitly maximizing its rank distance from the ground-truth class. Extensive results in various settings have verified that our proposed new loss is simple yet effective for top-$k$ attack. Code is available at: \url{https://bit.ly/3uCiomP}

\end{abstract}

\section{Introduction}
\label{Introduction}
Deep neural networks (DNNs) are widely known to be vulnerable to adversarial examples~\cite{huang2019black,ilyas2018black,wu2020decision,zhang2020understanding}, which are crafted by adding imperceptible or quasi-imperceptible perturbations to natural images. This intriguing phenomenon has inspired a vibrant field of studying the model robustness~\cite{xie2019intriguing,mummadi2019defending,shafahi2019adversarial,wang2019bilateral}. One intriguing property of adversarial examples is the widely known transferability from one (surrogate) model to another (target) model~\cite{hashemi2020transferable,li2019regional}. This property has been exploited for the transferable black-box attack as well as enhancing query-based black-box attack~\cite{inkawhich2020perturbing}. 

It is widely reported that I-FGSM increases the attack strength of FGSM, but at the cost of a lower transfer rate. This leads to a popular belief that the white-box strength of an attack is at odds with its transferability~\cite{kurakin2016adversarial}. Lower transfer rates of I-FGSM are often attributed to the conjecture that longer iterations lead to over-fitting to the surrogate model~\cite{kurakin2016adversarial,dong2018boosting}. Partly due to this concern, conventionally, existing works on transferable attack often adopt a limited number of iterations $T$, typically set to $\epsilon/\alpha$ where $\epsilon$ and $\alpha$ are the maximum $L_{\infty}$ budget and step size, respectively. In contrast, we show that this phenomenon can be at least partially explained by the lower perturbation magnitude of I-FGSM and a larger $T$ improves the transferability, eventually outperforming FGSM given a sufficiently large $T$. We further demonstrate that complementary to existing techniques, increasing $T$ consistently enhances the transferability and then saturates to a plateau.

Conventionally, attack success rate (ASR), also called fooling ratio (FR), is commonly used for evaluating strength in white-box, and transferability in black-box attacks. However, ASR does not provide an in-depth indication of attack strength. In essence, ASR only indicates whether an interest class, ground-truth class in the non-targeted or target class in targeted setting, ranks top-$1$ in the adversarial example. It would be interesting to know the ASR@$k$, \ie\ beyond from top-$1$ to top-$k$, to have a wide-range evaluation of attack strength. To this end, we introduce a new metric termed \textit{interest class rank (ICR)}, which facilitates the ASR@$k$ evaluation and, more importantly constitutes a single unified value indicating the \textit{top-$k$ attack strength}. 

With the ICR as the metric, we find that increasing $T$ enhances both top-$k$ adversarial strength and transferability, suggesting top-$k$ attack strength is also transferable. However, simply increasing the $T$ is not sufficient enough for a strong top-$k$ attack. We identify that the reason lies in the commonly used cross-entropy (CE) loss or C\&W loss which prioritize the speed of fooling the network instead of maximizing its distance from the interest (ground-truth) class. To this end, we propose Relative Cross-Entropy (RCE) loss for boosting stronger top-$k$ attack. Our new loss achieves close-to-optimal top-$k$ strength in white-box attack, outperforming existing losses by a large margin, consequently leading to a stronger top-$k$ transferable attack.

\textbf{Contributions.} Our work is the first to attempt the task of \textit{top-$k$ transferable attack}. A major obstacle towards this task is a popular belief on strength and transferability, which we challenge by empirical finding that top-$k$ attack strength is transferable. We identify the limitation of existing losses and propose a new RCE loss for achieving a strong top-$k$ attack in both white-box and transferable black-box settings. We extensively validate its efficacy for benchmarking top-$k$ strength and transferability of adversarial examples on multiple datasets. Even though our RCE loss is motivated to increase attack strength in the non-targeted setting, we show that it can also be easily extended to the targeted setting for boosting the transferability, especially for ASR@$1$. 

\section{Related work}
\textbf{Beyond attack success rate.} Although attack success rate (ASR) is a popular metric for evaluating the attack strength, its limitation comes in that it only shows whether the interest class, ground-truth class in the typical non-targeted setting, ranks top-1 after an attack. This limitation has been first noted in~\cite{kurakin2016physical}. To this end, ASR@$k$ (with a different term) has been introduced in~\cite{mopuri2020adversarial} has been introduced. For a given $k$, an attack is successful if the rank of the ground-truth class is larger than or equal to $k$. When $k$ is larger than 1, the attack is strictly more difficult than adopting the conventional ASR, \ie\ ASR@1, as the metric. In other words, if an attack is successful under ASR@$k$, it is guaranteed that it is a successful attack with the conventional ASR, but not vice versa. Extended to the targeted setting, an attack is successful if the rank of interest class, \ie\ target class, is smaller than the given $k$. When $k$ increases, task complexity of non-targeted and targeted settings increases and decreases, respectively.~\cite{ganeshan2019fda} has also proposed alternative metrics, such as old label new ranking (OLNR), new label old ranking (NLOR), cosine similarity (CosSim), normalized rank transformation (NRT). Complementary to their metrics, our work introduces a straightforward metric, interest class rank (ICR), to indicate the rank of the interest class after the attack. A major merit of ICR is that it can be directly transformed to ASR@$k$ for any $k$.

\textbf{Transferability and Black-box Attacks.} Various works have attempted to explain transferability from different perspectives. For example,~\cite{goodfellow2014explaining} attributes it to the hypothesis on the linear nature of modern DNNs, which has been recently supported by the recent finding that backpropagating with more wieght on skip connections improves transferability~\cite{wu2020skip}. Understanding transferability from the perspective of pixel interaction~\cite{wang2021unified} has also been investigated. Through the lens of non-robust feature~\cite{ilyas2019adversarial}, a recent work~\cite{benz2021batch} has shown that adversarial tranferability can be improved by removing BN from the surrogate model. Even though the rationality behind transferability is still not fully understood, this intriguing property has been widely exploited for black-box attacks. Early works have shown that adversarial examples naively generated in the direct white-box manner, such as vanilla I-FGSM, have low transferability. An ensemble of multiple surrogate models is found to improve the transferability~\cite{liu2016delving,tramer2017ensemble} but at the cost of more computation resources. Some free techniques have been proposed, such as momentum update~\cite{dong2018boosting}, input diversity~\cite{xie2019improving}, and translation-invariant constraint~\cite{dong2019evading}. \cite{huang2019enhancing,li2020yet} have demonstrated that fine-tuning adversarial examples with the intermediate level attack can further boost the transferability. Backpropagating linearly~\cite{guo2020backpropagating} or smoothly~\cite{zhang2021backpropagating} is also found to improve trasnferability.
Most investigation on transferable attack centers around non-targeted setting, and recently multiple works~\cite{inkawhich2019feature,inkawhich2020transferable,inkawhich2020perturbing} have attempted the targeted setting via the loss optimization in feature space. This often requires training additional class-wise layer-wise auxiliary classifiers. Directly performing the loss optimization in the output space is more simple but often at the loss of lower targeted transferability. Identifying the gradient vanishing issue of the cross-entropy (CE) loss,~\cite{li2020towards} has proposed a new loss based the Poincar\'{e} ball distance. %

\textbf{Positioning our work.} Our work is the first to propose a new task, namely top-$k$ transferable attack which aims to increase ASR-$k$ for a wide range of $k$ including $k=1$. For the top-$k$ attack, prior works exclusively only study in the white-box setting. On the other hand, prior works that study transferability do not take top-$k$ into account. Through showing attack strength is transferable, our work aims to realize top-$k$ transferable attack by increasing its white-box attack strength. Prior techniques improve adversarial transferability mainly through a regularization effect. Our investigated direction can be seen as orthogonal and complementary to them, which is supported by our empirical results. 

\section{Background and a popular belief} 
\textbf{White-box attacks.} White-box attack methods typically assume that the attacker has full knowledge of a target model, \ie\ the architecture and parameters\cite{carlini2017towards,madry2017towards,szegedy2013intriguing}. To make the adversarial perturbation imperceptible, the perturbation is often constrained inside a certain allowable perturbation budget or its $L_p$ is smaller than a certain magnitude, \ie\ $||v||_p \leq \epsilon$~\cite{szegedy2013intriguing}. Under such constraint, the goal of most existing adversarial attacks is to maximize a certain loss $L(x+v, y)$ for which the CE loss is widely used. 

\textbf{FGSM.} Goodfellow \etal proposed FGSM to craft adversarial examples: $X^{adv} = X + \epsilon sign(\nabla_X J(X, y_{true}))$, where $X^{adv}$ is the resulting adversarial example, $X$ is the attacked image, $J$ is the loss, $y_{true}$ is the ground truth label, and $\epsilon$ is the maximum allowable perturbation budget for making the resulting adversarial example look natural to the human eye. Simple FGSM achieves a reasonably high ASR. 

\textbf{Single-Step Least Likely Class Method (Step-LL).} This attack can be considered as a new variant of FGSM with a loss that targets a non-ground truth class~\cite{kurakin2016physical}: $X^{adv} = X + \epsilon \textnormal{sign} (\nabla_X J(X, y_{LL}))$, where $y_{LL} = \textnormal{arg min} (h(X))$, indicating the least-likely (LL) class based on the model output, \ie\ logit vector $h(X)$.

\textbf{I-FGSM or Iter-LL.} Iterative attack  was introduced in~\cite{kurakin2016physical,kurakin2016adversarial} to increase ASR by iteratively applying FGSM or Step-LL with the step size $\alpha$: $X^{adv}_{0} = X, X^{adv}_{t+1} = X^{adv}_{t}  + \alpha sign(\nabla_X J(X^{adv}_t, y))$. The step size $\alpha$ is often set to $\epsilon / T$, where $T$ indicates the number of iterations, for satisfying the $L_{\infty}$ constraint. It has been widely reported in~\cite{kurakin2016physical,kurakin2016adversarial,dong2018boosting} that iterative attack methods induce a higher ASR than their single-step counterparts, \ie~ FGSM or step-LL, but \textit{transfer} at lower success rates. For example, it is argued in~\cite{kurakin2016adversarial} that \textit{``there might be an inverse relationship between transferability of specific method and ability of the method to fool the network,''} which implies that adversarial strength is at odds with transferability. 

\begin{figure*}[t]
    \centering
    \includegraphics[width=\linewidth]{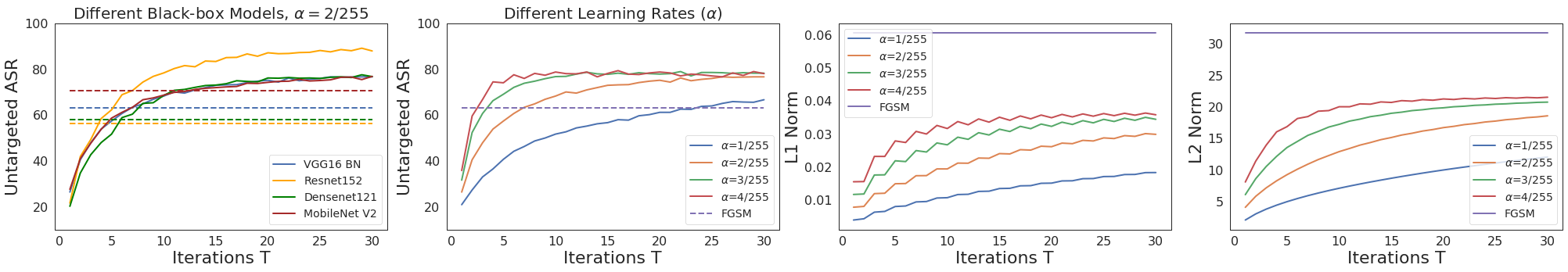}
    \caption{Transferability result for the FGSM (dashed lines) and I-FGSM (solid lines) with source network ResNet50 (RN50) and various black-box models (1st left). Performance for different step sizes ($\alpha$) when black-box model is VGG16 (2nd left). L1 (3rd left) and L2 (4th left) norms of the perturbation over the iterations. L1 norm is calculated on all pixel dimensions as an averarage their absolute values.}
    \label{fig:transferability_with_diff_step_sizes}
\end{figure*}

\begin{figure*}[!htbp]
    \centering
    \includegraphics[width=\linewidth]{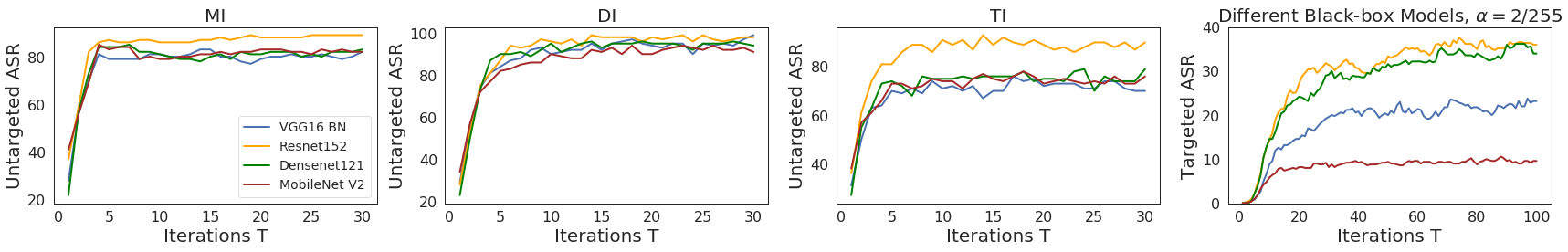}
    \caption{\textbf{First three figures from the left:} Non-targeted transferability with MI, DI, and TI. \textbf{Rightmost figure:} Targeted transferability with the MI-DI-TI-FGSM. The source network is ResNet50. 
    } 
    \label{fig:transferability_with_mi_di_ti}
\end{figure*}

\textbf{Existing techniques for improving transferability.} Most techniques introduced in popular works for improving transferability play the role of regularization.
It has been shown in~\cite{zhou2018transferable} that adding a regularization term can non-trivially improve the transferability. This is conceptually analogous to the practice of regularizing model training to avoid over-fitting, \ie\ slightly reducing the training accuracy, for improving the test accuracy. Other works have also introduced other implicit regularization techniques, such as gradient update with momentum~\cite{dong2018boosting}:
\begin{equation}
\begin{aligned}
g^{adv}_{t+1} = \mu g^{adv}_{t} + \frac{\nabla_X J(X^{adv}_t, y)}{||\nabla_X J(X^{adv}_t, y)x||_{1}}, \\
X^{adv}_{t+1} = X^{adv}_{t}  + \alpha sign(g^{adv}_{t+1}).
\end{aligned}
\end{equation}
where $\mu$ indicates the momentum weight, usually set to 1. The above technique is often called MI-FGSM. Another two famous variants of I-FGSM are DI-FGSM introduced in~\cite{xie2019improving} and TI-FGSM in~\cite{dong2019evading}. The DI-FGSM is shown as :
\begin{equation}
X^{adv}_{t+1} = X^{adv}_{t}  + \alpha sign(\nabla_X J(Tr(X^{adv}_t;p), y))
\end{equation}
where $Tr$ indicates transformation with the probability $p$. The TI-FGSM is shown as:
\begin{equation}
X^{adv}_{t+1} = X^{adv}_{t}  + \alpha sign(W * \nabla_X J(X^{adv}_t, y))
\end{equation}
where $W$ is a kernel for smoothing the gradients.

\textbf{Experimental Setup.}
Following previous works~\cite{dong2018boosting,dong2019evading,li2020towards}, we evaluate our proposed techniques on an ImageNet-compatible dataset composed of $1000$ images. This dataset was introduced in the NeurIPS 2017 adversarial challenge\footnote{\url{https://github.com/rwightman/pytorch-nips2017-adversarial}} and widely used for transferable black-box attack. 
Consistent with previous methods, we set the maximum perturbation magnitude to $L_\infty = 16/255$. 

\textbf{Influence of $\alpha$ and $T$ on transfer rate.} 
The phenomenon that FGSM is more transferable is often attributed to the fact that iterative attack methods tend to over-fit to the surrogate model~\cite{dong2018boosting,kurakin2016adversarial}. However, it remains yet unclear which factor mainly contributes to over-fitting. Technically, the differences between I-FGSM and FGSM  consist of two factors: step size $\alpha$ and number of iteration $T$. To demystify this, we analyze the influences of $\alpha$ and $T$ on the transfer rate. The results are shown in Figure~\ref{fig:transferability_with_diff_step_sizes}. We have two major observations: (a) Given a fixed $\alpha$, increasing $T$ enhances the transfer rate; (b) Given a fixed $T$, increasing $\alpha$ significantly boosts the transfer rate, especially when $T$ is not sufficiently large. The results demonstrate that \textit{the factor that contributes to the over-fitting of I-FGSM is its small $\alpha$ rather than its large $T$}. We find that given sufficiently large iterations, I-FGSM transfers better than FGSM. We report similar phenomenon in different setups in Figure~\ref{fig:transferability_with_mi_di_ti}. 

\textbf{Correlation with the $L_2$ norm.} Why does increasing $\alpha$ and $T$ enhance the transferability? We identify $L_2$ norm of the perturbation as a factor that correlates with the transferability. The results in Figure~\ref{fig:transferability_with_diff_step_sizes} show that in the setup of I-FGSM, there is a high positive correlation between transfer rate and $L_2$ norm (similar trend is observed for $L_1$ norm). Nonetheless, $L_2$ is not the only influence factor, otherwise, I-FGSM can never outperform FGSM for transferability.

\begin{figure*}[t]
    \centering
    \scalebox{0.99}{
    \includegraphics[width=\linewidth]{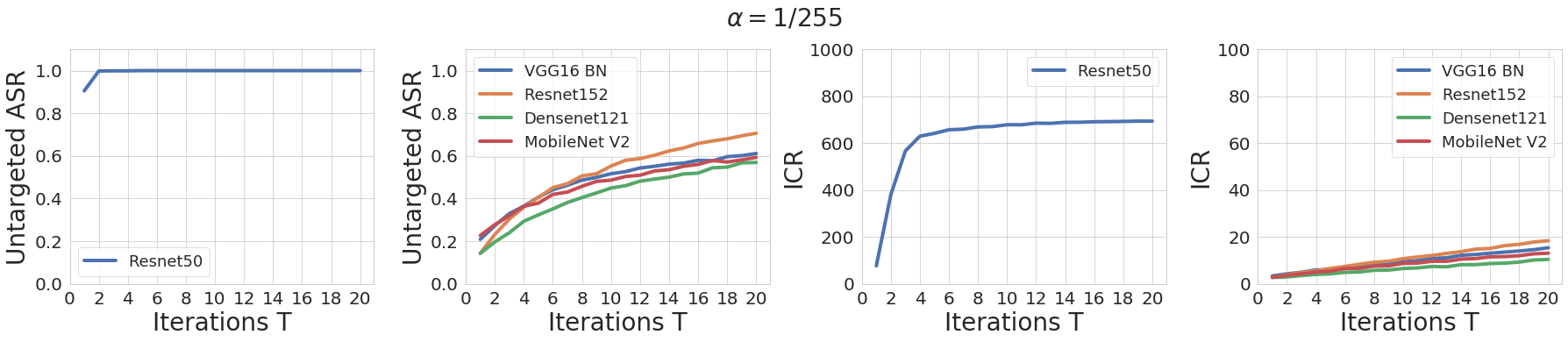}}
    \scalebox{0.99}{
    \includegraphics[width=\linewidth]{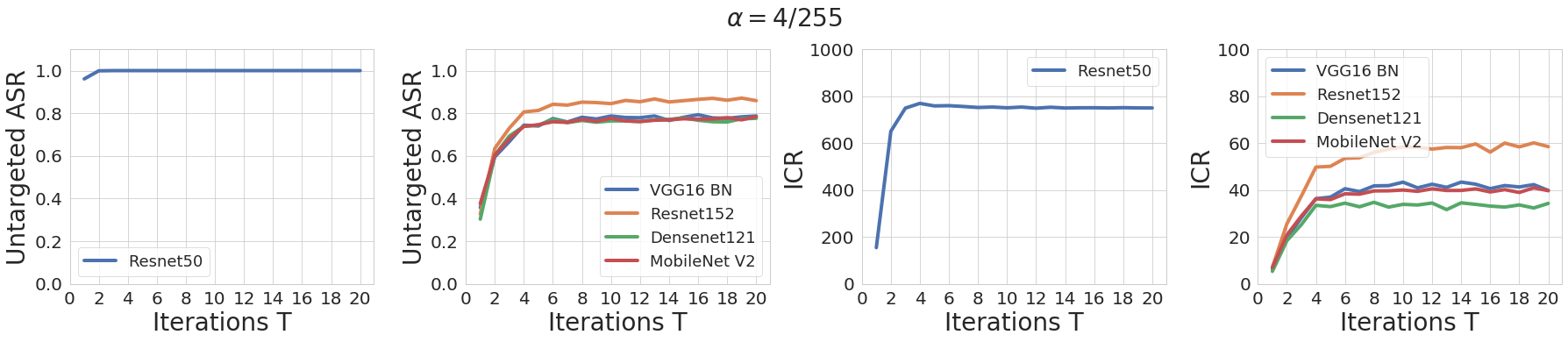}
    }
    \caption{ICR and ASR with $\alpha$ set to 1/255 (top) and 4/255 (bottom) with ResNet50 as the surrogate (white-box) model.} 
    \label{fig:reg_non_reg_icr_over_t_iterations_vis}
\end{figure*}

\section{Top-$k$ attack and ICR metric} 
\label{sec:adv_strength_transferability}

\textbf{Interest class rank.} With the top-$k$ metric, \ie\ ASR@$k$, there is no end to what $k$ can be, thus alternatively we also introduce a new metric called Interest Class Rank (ICR) which directly indicates the rank of the interest class after the attack. Note that for any sample, given the ICR value, we can easily tell whether it is a successful attack at any given $k$. For example, in an untargeted setting, if the ICR is 20, the attack is successful when $k$ in ASR@k is set to 10 ($10 < ICR$) and unsuccessful when $k$ is 30 ($30 > ICR$). Thus, without the need to enumerate all $k$, the ICR with a single value indicates the full-spectrum top-$k$ attack strength. Note that ICR can be used for both attack settings, where a larger ICR indicates the attack is \textit{stronger} in the untargeted setting and \textit{weaker} in the targeted setting. We highlight that the ICR is equivalent to ASR@$k$, since ICR can be easily transformed top-$k$ for any $k$.

\textbf{Top-$k$ attack strength is transferable.} With the ICR as the metric, we study the new rank between the surrogate model and target model, \ie\ whether the top-$k$ adversarial strength is transferable. Through analyzing a single sample, we observe that a higher ICR on the surrogate model also leads to a higher ICR on the target model, suggesting top-$k$ adversarial strength is transferable. Averaging on 1000 samples, we show the ICR with different $\alpha$ and $T$ and the results are shown in Figure~\ref{fig:reg_non_reg_icr_over_t_iterations_vis}. As a control study, we also report the same results with the metric of ASR-$1$. 
The overall trend of the ICR mirrors that of ASR. For example, either increasing $T$ or $\alpha$ significantly boosts the top-$k$ adversarial strength on the target model. However, it is more challenging to get satisfactory performance with the ICR metric. With $\alpha$ set to 1/255, even after 20 iterations, the black-box average ICR is only around 15/1000 (Maximum $K$ is 1000 for ImageNet). Adopting a larger $\alpha$ boosts convergence, however, the final ICR is still only around 40/1000. An important takeaway from the above results is that \textit{strong top-$k$ black-box attack might be achievable through increasing the white-box top-$k$ attack strength}.

\section{Boosting top-$k$ white-box attack}
\textbf{One intriguing property of logit vectors.} Let the logit vector be defined as the pre-softmax output of a DNN classifier and be denoted as $\mathbf{Z}$. Here, we report that the sum of all values in the $\mathbf{Z}$ vector is very close to zero in the vast majority of cases. We confirmed this phenomenon over various networks on different datasets for both adversarial examples and natural examples. Refer to the supplementary for detailed results of this intriguing phenomenon as well as a possible explanation. Moreover, the zero-sum phenomenon indicates that the logit values in $\mathbf{Z}$ have to be internally connected to satisfy \textit{zero-sum} constraint. In the following, we present a geometric interpretation of the gradient directions of different loss functions, for which the zero-sum property of $\mathbf{Z}$ will constitute an important assumption.

\textbf{Gradient directions of common loss functions.}
The influence of the loss on the generation of adversarial examples lies in the perturbation gradient update direction. Due to the extremely non-linear behavior of the network, it is intractable to intuitively derive a loss by analyzing the gradient on the network input. To alleviate such concern, we focus on tractable gradients of the logit vector. In other words, we assume that we can directly update the logits. Admittedly, we recognize that directly updating logit is not practical since we can only update the input perturbation. Nevertheless, with the backward-propagation chain rule, an optimal gradient update on the logit will lead to a pseudo-optimal update on the input perturbation. In this part we will first discuss the gradient directions with respect to the logit vector $\mathbf{Z}$ of commonly used loss functions. The detailed derivations can be found in the supplementary and here we present the main results. For the non-targeted setting, the derivative with respect to $\mathbf{Z}$ for CE, CE(LL) and CW losses are $\mathbf{P} - \mathbf{Y}_{gt}$, $\mathbf{Y}_{LL} - \mathbf{P} $ and $\mathbf{Y}_{j} - \mathbf{Y}_{gt}$, respectively. $\mathbf{P}$ is the post-softmax probability vector and  $\mathbf{Y}_{gt}$, $\mathbf{Y}_{LL}$, $\mathbf{Y}_{j}$ ($j = \arg \max\limits_{i \neq gt} Z(X^{adv})_{i} $) indicate the ground-truth one-hot label, that of the least likely class, and that of highest class except the ground-truth class, respectively. 

\textbf{Relative CE Loss.}
Next, we present our new loss formulation for boosting the top-$k$  strength. The loss function, which we term Relative CE loss or $RCE$ in short is formulated as follows:
\begin{equation}
\resizebox{\hsize}{!}{
    $RCE(X^{adv}_t, y_{gt}) = CE({X^{adv}_{t}, y_{gt})} - \frac{1}{K}\sum_{k=1}^{K} CE({X^{adv}_{t}, y_{k})},$
    }
\end{equation}
which consists of two parts, the commonly used CE, and a normalization part, averaging the CE calculated for each class. Its gradient on the logit vector $\textbf{Z}$ is derived as follows:
\begin{equation}
\frac{\partial L_{RCE}}{\partial \mathbf{Z}} = \frac{1}{K} - \mathbf{Y}_{gt}.
\end{equation}

\begin{figure}[!htbp]
    \centering
    \includegraphics[width=0.99\linewidth]{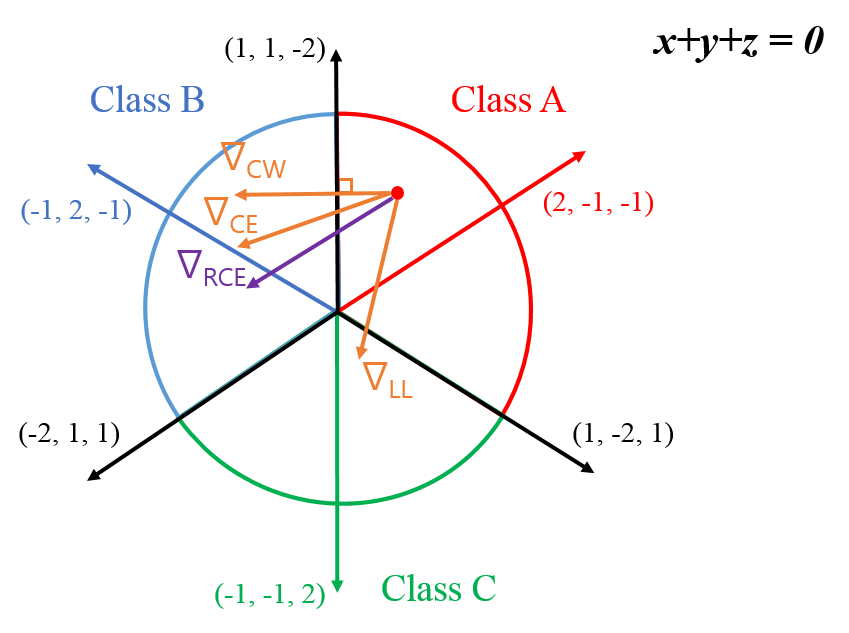}
    \caption{Geometric interpretation of the logit gradient of losses.}
    \label{fig:geom_vis}
\end{figure}

After establishing the gradient directions of common loss functions and the introduction of our loss function and its corresponding gradient, we provide a geometric perspective to illustrate why the proposed loss increases top-$k$ adversarial strength. In short, we will show that \textit{the gradient direction of the $RCE$ loss pushes a sample most far away from its ground-truth class.}

\textbf{Geometric interpretation of the logit gradient.} For illustration purpose, our setup is designed to have only three classes A, B, C. Each class is represented by the corresponding logit value $x$, $y$, and $z$, respectively. First, we assume that there is no constraint on the logits, thus each logit is fully independent. The logit space can be represented in the 3-D space with three orthogonal axes $X$, $Y$, and $Z$. Previously, we described the zero-sum phenomenon of logit vector $\mathbf{Z}$ that the sum of logits is always very close to zero for clean samples and adversarial samples. The logits are constrained to lie on a plane of $x + y + z = 0$ (with a normal vector of $(1,1,1)$), which is termed (logit) decision hyperplane. In other words, the \textit{zero sum} constraint decreases the degree of freedom from 3-D space to a 2-D plane. We visualize this 2-D plane in Figure~\ref{fig:geom_vis}. 

With the symmetric assumption, the direction of the class-wise logit vector for class A, B, C can be set to $(2,-1,-1)$, $(-1,2,-1)$, $(-1,-1,2)$ with a certain scale. We highlight that vector scale is irrelevant and only the direction matters due to the sign function on the input gradient processing, \ie\ FGSM. It is worth mentioning that the sum of the values in the $\frac{\partial L}{\partial \textbf{Z}}$ is also always equal to zero for the above discussed three losses. Moreover, all the points on the plane satisfy $x + y + z = 0$ given the \textit{zero sum} constraint. Thus, all the discussion here is always on the decision hyperplane $x + y + z = 0$. Suppose, at step $t$, the position of the sample on the decision hyperplane is $(x_{t},y_{t},z_{t})$. Without losing generality, we assume the sample is on the region of class $A$ and $y_{t} > z_{t}$ indicating the sample is more close to the logit decision boundary with $B$ than $C$. 

To give a concrete example for facilitating the discussion, we assume $x_t = 1, y_t = 0.2, z_t = -1.2$ and the resulting post-softmax probability vector is $P=(0.64, 0.29, 0.07)$. We assume that the sample is correctly classified, hence its ground truth vector is $Y_{gt} = (1,0,0)$. Following the descriptions above $Y_{LL}=(0,0,1)$ $Y_j=(0,1,0)$, the calculated derivatives for CE, CW and CE(LL) are detailed as:
\begin{equation*}
\resizebox{1\hsize}{!}{
$\frac{\partial L_{CE}}{\partial \textbf{Z}} = \begin{pmatrix}-0.36 \\ 0.29\\ 0.07\end{pmatrix}; \frac{\partial L_{CE(LL)}}{\partial \textbf{Z}} = \begin{pmatrix} -0.64 \\ -0.29 \\ 0.93 \end{pmatrix};
\frac{\partial L_{CW}}{\partial \textbf{Z}} = \begin{pmatrix} -1\\ 1\\0 \end{pmatrix}; \frac{\partial L_{RCE}}{\partial \textbf{Z}} = \begin{pmatrix}-0.66 \\ 0.33\\ 0.33\end{pmatrix}$.}
\end{equation*}
With the gradient derivation, we find that CW and CE shift the sample towards class B while the CE(LL) shifts the samples to class C. A detailed comparison shows that the CW gradient direction is orthogonal to the decision boundary between A and B in this 3-class setup. Thus intuitively, CW loss prefers to encourage the sample to find the nearest decision boundary to cross. CE also results in a gradient direction that is close to the decision boundary of B. By contrast, our RCE loss does not explicitly encourage the sample to choose any decision boundary. All CW, CE, and CE(LL) share one common property: the logit update direction is dependent on the current sample position on the decision hyperplane. Depending on the position of the sample on the decision plane, CE and CW tend to move the sample towards a semantically close class (B in this case), while CE(LL) loss explicitly moves the sample to a semantically far class (C in this case). With the interest class as A, intuitively, a strong attack should maximize its semantic distance from class A, \ie\ updating in a direction that is opposite of the interest class logit vector. The gradient of our RCE loss adopts this direction regardless of the sample position on the decision hyperplane to move the sample far from class A. Due to ignorance of the current sample position, one drawback of our approach is that it might lead to relatively slower convergence. Empirically, we confirm that this is a concern in the very early iterations, see the supplementary for relevant discussion.

\begin{table}[!htbp]
\caption{Comparison of RCE loss with other losses in the white box scenario. The discrepancy between ICR and OLNR exists because not all samples in the dataset are correctly classified. }
\centering
\scalebox{0.70}{
\begin{tabular}{c|ccccccccc}
\toprule
& non-targeted Acc. & ICR & OLNR & NLOR & NRT & CosSim \\
\midrule
CE & 100.00 & 752.90 & 712.35 & 159.52 & 279.53 & 0.25 \\
CW & 100.00 & 391.40 & 349.94 & 21.01 & 257.22 & 0.40 \\ 
LL & 99.20 & 491.02 & 490.46 & 888.96 & 306.12 & 0.08 \\ 
FDA & 100.00 & 619.90 & 608.84 & 517.28 & 311.49 & 0.06 \\ 
\midrule
RCE(Ours) & 100.00 & \textbf{1000.00} & \textbf{979.63} & 570.94 & \textbf{360.23} & \textbf{-0.21} \\ 
RCE(LL) & 100.00 & 687.36 & 688.72 & \textbf{996.32} & 354.58 & -0.17 \\
\bottomrule
\end{tabular}
}
\label{tab:whitebox}
\end{table}

\textbf{Strong top-$k$ white-box attack.} Here, we compare our loss with CE, CW, CE(LL), and FDA~\cite{ganeshan2019fda}. The results are shown in Table~\ref{tab:whitebox}. Except for our proposed ICR metric, for completeness we also report other metrics in~\cite{ganeshan2019fda}, including OLNR, NLOR, cosine similarity (CosSim), normalized rank transformation (NRT), and ASR, for evaluating top-$k$ adversarial strength. The $\alpha$ and $T$ are set to $4/255$ and $20$ (same for other experiments, unless specified). The results show that our loss achieves the strongest attack among all losses for all metrics except for NLOR with CE(LL). Note that CE(LL) loss explicitly targets the LL class, thus it is expected NLOR would be higher. We further conduct an experiment with RCE(LL) which achieves $996.32$ for NLOR, significantly outperforming CE(LL).

\begin{table}[!htbp]
\caption{ICR under image transformations for different loss functions.}
\centering
\scalebox{0.77}{
\begin{tabular}{c|ccccccccc}
\toprule
& No transform & Brightness & Contrast & Gaussian Noise \\
\midrule
CW & 390.00 & 216.27 & 185.01 & 33.18 \\
CE & 752.90 & 488.92 & 460.19 & 71.28 \\ 
RCE (Ours) & 1000.00 & 897.85 & 876.94 & 201.25 \\ 
\bottomrule
\end{tabular}
}
\label{tab:transformation}
\end{table}

\textbf{Stronger top-$k$ attack under image transformations.} Following~\cite{kurakin2016physical}, we apply image transformations to the generated adversarial examples to test whether our loss still achieves stronger attack under image transformation (see Table~\ref{tab:transformation}). Note that such a setup constitutes testing the robustness of adversarial examples. Please refer to the supplementary for a detailed experimental setup.

\textbf{Loss comparison through the lens of temperature.}
From Figure~\ref{fig:geom_vis}, we observe that CE gradient direction lies between that of CW and our loss. Table~\ref{tab:whitebox} also shows that the performance of CE also lies in between. Here, we show that CW and our loss can be seen as a special case of CE through changing the temperature $T_e$~\cite{hinton2015distilling}. 
 
$T_e$ is a non-trivial hyperparameter temperature, \ie\ pre-processing to $\mathbf{Z} = \mathbf{Z} / T_e$ as the softmax input, resulting in $\mathbf{P}_e$. This temperature scaling method has been widely used for knowledge distillation~\cite{hinton2015distilling,cho2019efficacy} as well as a defense method~\cite{papernot2016distillation}. With the temperature taken into account, the derivative of the CE is derived as follows: 
\begin{equation}
\frac{\partial L}{\partial \textbf{Z}} = \frac{1}{T_e} (\mathbf{P}_e - \mathbf{Y}_{gt}), 
\end{equation}
Typically, the temperature $T_{e}$ is set to 1. From our geometric perspective, the $T_{e}$ balances the preference of the loss to encourage the sample to cross the decision boundary of the semantically closer class, \ie\ those classes with relatively high logits. A higher $T_{e}$ indicates decrease of such preference. With the temperature $T_{e}$ as a control variable, we reveal that the CW loss can be interpreted as a special case of the CE loss by setting $T_{e}$ to a small value. Our proposed $RCE$ loss can also be seen as a special case of the CE loss by setting $T_{e}$ to a sufficiently large value. The proof is given in the supplementary. Empirically, we demonstrate the influence of temperature on the attack strength in Table~\ref{tab:transferability_temperature}. The results validate that increasing/decreasing the $T_e$ shifts the performance close to RCE/CW loss.

\begin{table*}[!htbp]
\caption{Influence of different temperature values in the CE loss. Metric adopted is ICR.}
\centering
\scalebox{0.99}{
\begin{tabular}{cccccccccccc}
\toprule
CW & $T_{e}=1/100$ & $T_{e}=1/8$ & $T_{e}=1/4$ & $T_{e}=1/2$ & $T_{e}=1$ & $T_{e}=2$ & $T_{e}=4$ & $T_{e}=8$ & $RCE$ \\
\midrule
55.53 & 76.89 & 346.74 & 393.98 & 491.71 & 752.90 & 947.48 & 987.93 & 999.60 & 1000.0 \\
\bottomrule
\end{tabular}
}
\label{tab:transferability_temperature}
\end{table*}

\begin{figure*}[!htbp]
    \centering
    \includegraphics[width=1\linewidth]{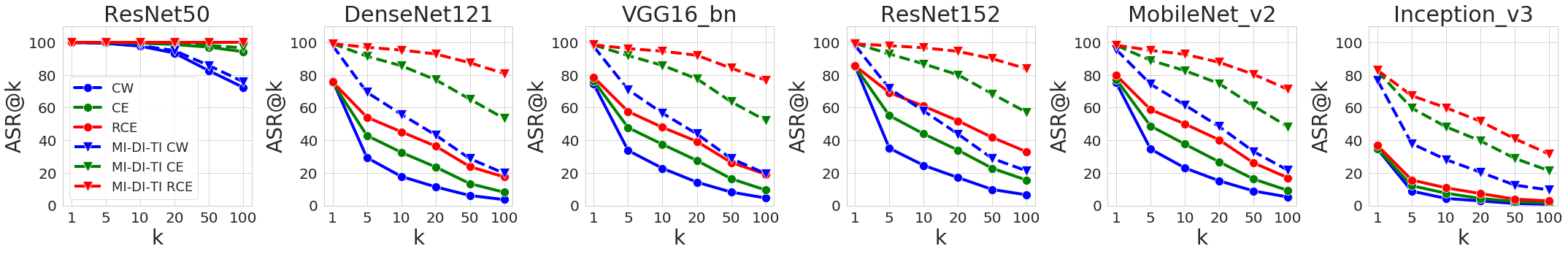}
    \includegraphics[width=1\linewidth]{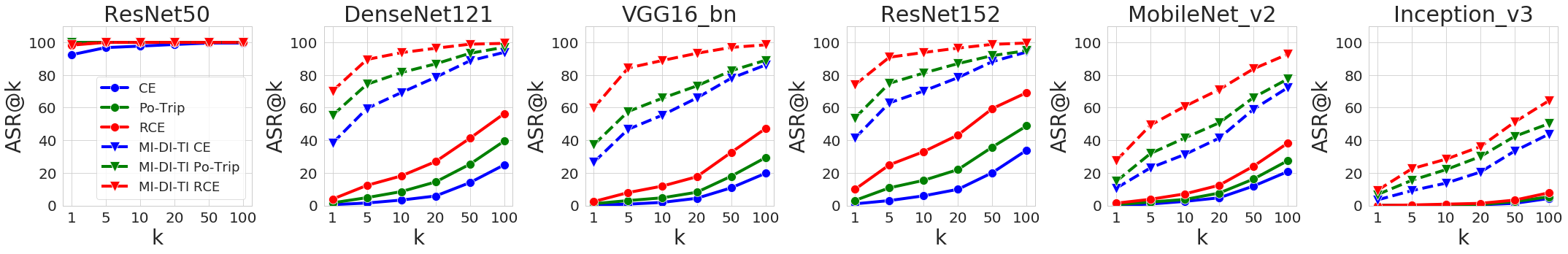}
    \caption{ASR@$k$ transferability with ResNet50 as the surrogate model in untargeted setting (top row) and targeted setting (bottom row)} 
    \label{fig:untargeted_targeted_fr_k_with_diff_models}
\end{figure*}

\section{Proposed top-$k$ transferable attack}\label{sec:transferable_strong}
Inspired by the finding that top-$k$ adversarial strength is transferable, we believe that our loss might also lead to a stronger top-$k$ transferable attack since it achieves the strongest white-box attack. Unless specified, we always adopt $\alpha=4/255$. We set $T$ to 20 and 200 for the non-targeted setting and targeted setting, respectively. 

\begin{table}[t]
\caption{Non-targeted transferability of I-FGSM (top), and MI-DI-TI-FGSM (bottom) attacks for the source network ResNet50. Each entry represents the ICR/non-targeted ASR@1 (\%).}
\centering
\scalebox{0.52}{
\begin{tabular}{cc|ccccccccc}
\toprule
 & RN50 & DN121 & VGG16bn & RN152 & MNv2 & IncV3 \\
\midrule
CW & 390.00/100.00 & 14.80/76.50 & 18.59/74.30 & 24.15/85.60 & 22.68/75.20 & 5.49/34.60 \\ 
CE & 752.90/100.00 & 34.16/75.40 & 40.87/76.40 & 61.20/85.20 & 39.21/77.30 & 7.50/34.80 \\ 
RCE (Ours) & 1000.00/100.00 & 72.11/75.80 & 80.86/78.50 & 144.81/85.60 & 70.39/79.80 & 13.35/36.80 \\ 

\midrule
CW & 427.49/100.00 & 77.82/98.10 & 77.13/97.40 & 81.67/98.20 & 84.88/95.60 & 39.03/76.80 \\ 
CE & 806.85/100.00 & 220.87/99.30 & 213.77/98.40 & 249.02/99.40 & 193.96/98.20 & 89.93/82.40 \\ 
RCE (Ours) & 999.94/100.00 & 482.58/99.20 & 430.97/98.50 & 517.85/99.00 & 366.30/98.30 & 141.90/83.00 \\ 

\bottomrule
\end{tabular}
}
\label{tab:non-targeted_transferability}
\end{table}

\textbf{Non-targeted setting.} The results are shown in Table~\ref{tab:non-targeted_transferability}, where we compare our loss with CE and CW in two different baselines: vanilla I-FGSM and MI-DI-TI-FGSM. For both baselines, our RCE loss outperforms CE loss by a large margin. The top row in Figure~\ref{fig:untargeted_targeted_fr_k_with_diff_models} shows that our loss also achieves higher (untargeted) ASR@$k$, especially when $k$ is set to large for making the task more challenging.

\begin{table}[!htbp]
\caption{Targeted transferability of I-FGSM (Top), and MI-DI-TI-FGSM (bottom) attacks for the source network ResNet50. Each entry represents the ICR/targeted ASR@1 (\%).}
\centering
\scalebox{0.56}{
\begin{tabular}{c|ccccccccc}
\toprule
 & RN50 & DN121 & VGG16bn & RN152 & MNv2 & IncV3 \\
\midrule
CE & 2.52/92.40 & 320.73/0.50 & 355.33/0.30 & 264.20/1.00 & 345.40/0.00 & 607.46/0.00 \\ 
Po-Trip & 1.00/100.00 & 236.37/1.60 & 299.51/1.10 & 192.63/3.10 & 309.81/0.50 & 582.28/0.00 \\
RCE (Ours) & 1.02/98.30 & 161.13/3.90 & 208.61/2.40 & 108.22/9.90 & 244.40/1.40 & 559.95/0.00 \\
\midrule
CE & 1.00/100.00 & 22.19/38.20 & 45.64/26.50 & 23.61/41.30 & 92.72/10.60 & 245.79/3.40 \\ 
Po-Trip & 1.00/100.00 & 13.84/55.30 & 40.33/37.20 & 18.46/53.70 & 76.37/14.80 & 215.26/6.70 \\
RCE (Ours) & 1.01/98.90 & 4.51/70.20 & 7.76/59.80 & 3.67/74.00 & 30.90/27.50 & 157.35/9.30 \\ 
\bottomrule
\end{tabular}
}
\label{tab:targeted_transferability_resnet}
\end{table}

\begin{table}[!htbp]
\caption{Targeted transferability of I-FGSM (Top), and MI-DI-TI-FGSM (bottom) attacks for an ensemble of source networks ResNet50 and DenseNet121. Each entry represents the ICR/targeted ASR@1 (\%).}
\centering
\scalebox{0.55}{
\begin{tabular}{c|ccccccccc}
\toprule
 & RN50 & DN121 & VGG16bn & RN152 & MNv2 & IncV3 \\
\midrule
CE & 2.07/92.00 & 1.60/96.00 & 242.62/2.20 & 175.88/4.20 & 258.10/1.60 & 521.37/0.00 \\ 
Po-Trip & 1.00/99.90 & 1.00/100.00 & 203.21/5.30 & 130.12/11.00 & 230.62/2.10 & 492.40/0.40 \\
RCE (Ours) & 1.02/98.30 & 1.02/98.50 & 78.95/16.20 & 44.90/29.00 & 135.20/5.80 & 419.23/0.50 \\ 

\midrule
CE & 1.00/100.00 & 1.00/100.00 & 15.74/54.90 & 8.22/66.60 & 43.16/23.50 & 119.14/14.90 \\ 
Po-Trip & 1.00/100.00 & 1.00/100.00 & 27.86/48.70 & 11.08/65.50 & 53.26/24.20 & 136.38/14.90 \\
RCE (Ours) & 1.01/98.80 & 1.01/98.80 & 2.48/81.70 & 1.86/86.80 & 10.21/50.10 & 59.86/30.70 \\ 
\bottomrule
\end{tabular}
}
\label{tab:targeted_transferability_ens}
\end{table}

\textbf{Targeted setting.} Table~\ref{tab:targeted_transferability_resnet} shows the results in the targeted setting, where a smaller targeted ICR indicates superior performance. Our approach achieves significantly better performance than CE and Po-Trip loss~\cite{li2020towards} which constitutes the SOTA approach that generates perturbation in the output space. The bottom row in Figure~\ref{fig:untargeted_targeted_fr_k_with_diff_models} shows that our loss also results in a higher targeted ASR@$k$, especially when $k$ is set to 1. For example, from ResNet to VGG16, our loss improves the performance of Po-Trip loss from 37.20\% to 59.80\%. It also outperforms another approach that generates perturbation in the feature space~\cite{inkawhich2020perturbing} (59.80\% vs.\ 43.5\%). Moreover, our loss also achieves comparable targeted transferability as a recent work~\cite{zhao2021success} that adopts logit loss~\cite{zhang2020understanding} for only maximizing the logit for targeted UAP (UAP is perturbation that fools the model for most images~\cite{moosavi2017universal,benz2020double,zhang2021universal,zhang2021survey}). We further conduct experiments in the ensemble setting by generating adversarial examples on source networks ResNet50 and DenseNet121(see results in Table~\ref{tab:targeted_transferability_ens}). Following~\cite{benz2021robustness}, we also report the performance on ViTs~\cite{dosovitskiy2021an} and MLP-Mixer~\cite{tolstikhin2021mlpmixer} (See results in the supplementary.) We observe that our RCE loss consistently outperforms the existing losses by a significant margin.

\textbf{CIFAR results.} We further conduct experiment on CIFAR10 (see Table~\ref{tab:non-targeted_transferability_cifar10}) and CIFAR100 (see Table~\ref{tab:non-targeted_transferability_cifar100}). The trend mirrors that on the ImageNet in both non-targeted and targeted settings.

\begin{table}[!htbp]

\caption{Non-targeted (top) and targeted (bottom) ICR/ASR@1 of the MI-FGSM attack on CIFAR-10 with surrogate ResNet50.}
\centering
\scalebox{0.80}{
    \begin{tabular}{c|cccccccccc}
    \toprule
     & RN20 & RN56 & VGG19 & DN \\
    \midrule
    CW & 6.03/99.70 & 6.07/99.70 & 5.04/98.40 & 6.82/99.60 \\ 
    CE & 6.23/99.50 & 6.28/99.40 & 5.24/98.40 & 6.83/99.20 \\ 
    RCE (Ours) & 8.23/99.10 & 8.00/99.10 & 6.80/96.10 & 8.50/98.70 \\ 
    \midrule
    CE & 1.11/93.40 & 1.12/93.20 & 1.24/87.80 & 1.05/95.50 \\ 
    Po-Trip & 1.64/71.10 & 1.62/68.70 & 1.77/69.00 & 1.43/79.80 \\
    RCE (Ours) & 1.08/94.60 & 1.08/94.30 & 1.18/89.80 & 1.03/98.00 \\ 
    \bottomrule
    \end{tabular}
}
\label{tab:non-targeted_transferability_cifar10}
\end{table}

\begin{table}[!htbp]
\caption{Non-targeted (top) and targeted (bottom) ICR/ASR@1 of the MI-FGSM attack on CIFAR-100 with surrogate ResNet50.}
\centering
\scalebox{0.80}{
    \begin{tabular}{c|cccccccccc}
    \toprule
     & RN20 & RN56 & VGG19 & DN \\
    \midrule
    CW & 21.57/94.00 & 22.23/95.80 & 20.09/91.60 & 21.62/93.60 \\ 
    CE & 24.31/95.10 & 25.24/96.30 & 24.58/93.70 & 24.44/96.00 \\ 
    RCE (Ours) & 48.32/97.70 & 52.35/97.00 & 44.05/96.40 & 45.82/96.30 \\ 
    \midrule
    CE & 20.74/11.40 & 18.46/16.20 & 31.43/13.60 & 14.52/15.30 \\ 
    Po-Trip & 23.75/10.40 & 21.59/13.60 & 33.24/12.00 & 17.89/14.40 \\
    RCE (Ours) & 11.46/22.20 & 10.08/27.50 & 23.08/17.70 & 9.54/20.70 \\ 
    \bottomrule
    \end{tabular}
}
\label{tab:non-targeted_transferability_cifar100}
\end{table}

\textbf{Images with various types of content.} The goal of an attack against a deep classifier is to add a small perturbation for changing the model output. Here, we perceive this output change as shifting the sample \textit{far from} or \textit{close to} a certain interest class regardless of original image content. Our ICR can be used to indicate the attack strength in this general setting. Due to position-agnostic property, our RCE loss outperforms CE loss in all setups (see results in the supplementary). 

\section{Discussion}
\textbf{I-FGSM \textit{vs.} PGD.} I-FGSM~\cite{kurakin2016adversarial} and PGD~\cite{madry2017towards} are in essence the same except with a technical difference. Specifically, I-FGSM initializes the initial perturbation with zero values while PGD initializes it with random values. Random initialization of PGD allows multiple restarts if the attack fails. However, in the black-box setting, only a single attempt is allowed for the evaluation purpose, thus the community sticks to use I-FGSM based attack for transferable attack. That is why our experiments are also based on initialization-free I-FGSM. In the white-box setting, multiple starts are allowed, however, with a single run (no multiple starts), our loss already achieves 100\% ASR@$k$ even when $k$ is set to the maximum $K$.  

\textbf{Top-$k$ optimization vs.\ top-$k$ evaluation.}
~\cite{zhang2020learning} performs the ordered top-k targeted attack. With their definition, their attack considers attacking multiple classes at the same time. With this said, we emphasize that our work does \textit{not} preform a top-k otimization because our goal is to manipulate the rank of a \textit{single class}. Instead, our work only adopts the top-k \textit{metric} as the evaluation. In other words, the $k$ of ASR@$k$ is not utilized in the training stage; in evaluation, it is also recommended to report ASR@$k$ for a wide range of $k$ values. Moreover, our loss is \text{not} designed for directly maximizing the ICR (in the untargeted setting) which is a discrete thus non-differential optimization goal. With this said, our loss does not overfit to the ICR metric and improves the performance for all other existing metrics.

\textbf{Targeted vs.\ non-targeted top-$k$ attack.}  
Top-$k$ targeted attack is \textit{not} very meaningful if we only care about whether the prediction label after the attack is the target class. When the targeted attack goal fails, however, our ICR still conveys non-trivial information, \ie\ the rank of the target class. It is worth mentioning that non-targeted top-$k$ attack also has an \textit{implicit target} direction (far from the ground-truth class). With this said, we highlight that our ICR provides a unified perspective on targeted and non-targeted attack. For black-box transferability, we highlight that increasing ICR in the non-targeted setting is significantly more challenging than decreasing ICR in the targeted setting. In the scale of 1 to 1000 for ICR on ImageNet, the optimal performance is 1 in targeted setting and 1000 in non-targeted setting. Taking the trasnfer from ResNet50 to DesnseNet121 as example, our results in Table~\ref{tab:targeted_transferability_resnet} show that the best achieved targeted ICR is 4.51 which is very close to the optimal value of 1. On the other hand, the best achieved non-targeted ICR is 482.58(see Table~\ref{tab:non-targeted_transferability}) which is very far from the optimal value of 1000. Considering the saturated top-$1$ transferability, we advocate future works to evaluate their attack methods with our ICR, especially for the non-targeted setting. For evaluating the strength of transferable attacks, our ICR can be a more reliable and informative metric than top-$1$ ASR.

\section{Conclusion}
Our work is the first to investigate the top-$k$ transferable attack. Motivated from the finding that top-$k$ attack strength is transferable, we explore how to achieve strong top-$k$ white-box attack. With the limitation of existing losses identified as mainly prioritizing the speed to fool the network based on an intuitive geometric perspective, we propose a new RCE loss that conceptually maximizes its semantic distance from the ground-truth class. The proposed RCE loss achieves a significantly stronger top-$k$ white-box attack for a wide range of metrics, including our proposed ICR. Due to the transferable property of top-$k$ attack strength, our loss also achieves a top-$k$ transferable attack that is significantly stronger than the SOTA approaches. We further extend our loss to the targeted setting, where a significant performance boost is also observed.

{
        \small
    \bibliographystyle{ieee_fullname}
    \bibliography{macros,bib_mixed}
}

\newpage

\appendix

\setcounter{page}{1}

\twocolumn[
\centering
\Large
\textbf{[CVPR2022]} \\
\vspace{0.5em}Supplementary Material \\
\vspace{1.0em}
] \appendix

\section{Experimental Setup}
\subsection{General Setup}
Following previous works~\cite{dong2018boosting,dong2019evading,li2020towards}, we evaluate our proposed approach on an ImageNet-compatible dataset composed of $1000$ images. Consistent with prior works, we set the maximum perturbation magnitude to $L_\infty = 16/255$. The step size $\alpha$ is set to 4/255 and unless specified, we set the number of iterations $T$ to 20 and 200 for the non-targeted and targeted attack, respectively.

\subsection{Image Transformations}
To test the robustness of the generated adversarial examples to image transformations, we apply brightness, contrast, and Gaussian noise transformations. For the brightness and contrast transformations, we increase the brightness and contrast by a factor of $2$. For the Gaussian noise augmentation, we apply Gaussian noise centered around zero mean, with a standard deviation of $0.1$.

\section{Additional Results for supporting our proposed RCE loss}

\begin{table}[!htbp]
\caption{Results of transfering from CNN (ResNet50) to ViT and MLP backbones. Each entry represents the ICR/targeted ASR@1 (\%).}
\centering
\setlength{\columnsep}{0pt}\scalebox{0.60}{
\begin{tabular}{c|ccccccccc}
\toprule
& non-targeted Acc. & ICR & OLNR & NLOR & NRT & CosSim \\
\midrule
CE & 100.00 & 629.11 & 604.28 & 74.50 & 255.81 & 0.37 \\
CW & 100.00 & 276.59 & 258.51 & 5.28 & 237.44 & 0.48 \\
LL & 100.00 & 409.08 & 398.66 & 963.84 & 300.79 & 0.10 \\ 
FDA & 99.00 & 504.42 & 498.90 & 430.81 & 297.03 & 0.13 \\ 
\midrule
RCE(Ours) & 100.00 & 1000.00 & 984.15 & 477.37 & 346.03 & -0.15 \\ 
RCE(LL) & 100.00 & 567.13 & 555.13 & 996.23 & 342.56 & -0.12 \\
\bottomrule
\end{tabular}
}
\label{tab:whitebox_eps4}
\end{table}

\textbf{Smaller $\epsilon$ in white-box attack.} Table~\ref{tab:whitebox_eps4} reports the white-box results with an allowed $\epsilon$=4/255 on ResNet50. The trend mirrors that with $\epsilon$ set $16/255$ in the main manuscript.

\textbf{Transfer from CNN to ViT and MLP.} The results of transfering from CNN (ResNet50) to ViT and MLP are reported in Table~\ref{tab:targeted_transferability_vit_mlp}. We observe that our proposed RCE loss outperforms existing ones by a large margin.

\begin{table}[!htbp]
\caption{Results of transfering from CNN (ResNet50) to ViT and MLP. Each entry represents the ICR/targeted ASR@1 (\%). ResNet50 is trained by $l_2$-PGD atttack with $\epsilon$ set to 0.5.}
\centering
\scalebox{0.75}{
\begin{tabular}{cc|ccccccccc}
\toprule
&  & ViT B16 & ViT L16 & MLP-M B16 & MLP-M L16 \\
\midrule
\parbox[t]{1mm}{\multirow{3}{*}{\rotatebox[origin=c]{90}{I-FGSM}}}
& CE & 187.03/3.0 & 218.54/7.0 & 147.16/9.0 & 211.69/6.0  \\
& Po-Trip & 140.45/23.0 & 159.86/21.0 & 83.11/20.0 & 139.43/12.0 \\
& RCE & 48.53/28.0 & 66.58/25.0 & 43.98/31.0 & 71.57/17.0  \\
\midrule
\parbox[t]{1mm}{\multirow{3}{*}{\rotatebox[origin=c]{90}{MI-DI-TI}}}
& CE & 42.74/42.0 & 51.73/41.0 & 19.00/38.0 & 96.81/13.0 \\
& Po-Trip & 41.86/39.0 & 41.37/37.0 & 44.77/37.0 & 98.00/18.0 \\
& RCE & 23.35/49.0 & 32.67/54.0 & 10.90/44.0 & 46.62/28.0 \\
\bottomrule
\end{tabular}
}
\label{tab:targeted_transferability_vit_mlp}
\end{table}

\begin{table*}[!htbp]
\caption{Non-targeted transferability of I-FGSM (top), and MI-DI-TI-FGSM (bottom) attacks for source network DenseNet121. Each entry represents the ICR/non-targeted success rate (\%).}
\centering
\scalebox{0.85}{
\begin{tabular}{c|cccccccccc}
\toprule
 & RN50 & DN121 & VGG16bn & RN152 & MNv2 & IncV3 \\
\midrule
CW & 27.49/86.10 & 636.75/100.00 & 22.98/80.20 & 16.96/73.90 & 26.88/76.40 & 8.36/42.30 \\ 
CE & 68.10/86.70 & 851.94/100.00 & 58.14/84.90 & 39.74/75.30 & 51.90/79.20 & 14.22/45.70 \\ 
RCE (Ours) & 128.89/85.30 & 1000.00/100.00 & 100.56/83.50 & 72.46/75.10 & 85.67/82.90 & 19.32/44.10 \\ 

CW & 70.29/96.90 & 632.11/100.00 & 58.03/96.40 & 46.42/92.30 & 80.04/92.90 & 39.95/76.90 \\ 
CE & 206.55/98.50 & 883.52/100.00 & 192.35/97.80 & 138.87/95.50 & 168.07/96.70 & 99.17/84.30 \\ 
RCE (Ours) & 378.31/98.50 & 1000.00/100.00 & 337.00/98.10 & 254.97/95.20 & 288.82/97.50 & 144.69/82.80 \\ 

\bottomrule
\end{tabular}
}
\label{tab:non-targeted_transferability_ds121}
\end{table*}

\begin{table*}[!htbp]
\caption{Targeted transferability of I-FGSM (Top), and MI-DI-TI-FGSM (bottom) attacks for source network DenseNet121. Each entry represents the ICR/targeted success rate (\%).}
\centering
\scalebox{0.85}{
\begin{tabular}{c|ccccccccc}
\toprule
 & RN50 & DN121 & VGG16bn & RN152 & MNv2 & IncV3 \\
\midrule
CW & 195.81/4.60 & 1.11/99.90 & 219.33/2.60 & 265.05/1.20 & 277.73/0.70 & 533.24/0.10 \\ 
CE & 295.22/0.90 & 1.12/97.50 & 322.68/0.60 & 341.30/0.60 & 347.49/0.30 & 586.48/0.00 \\ 
RCE (Ours) & 154.12/5.20 & 1.01/98.70 & 175.77/4.00 & 226.36/1.70 & 254.80/0.80 & 510.23/0.30 \\ 
Po-Trip & 245.60/2.30 & 1.00/100.00 & 282.81/1.20 & 304.62/0.60 & 319.42/0.50 & 562.21/0.00 \\

\midrule
CW & 39.09/38.30 & 1.00/100.00 & 54.32/27.10 & 69.03/23.70 & 103.37/11.50 & 205.07/6.90 \\ 
CE & 79.21/15.70 & 1.00/100.00 & 107.42/10.90 & 118.32/7.80 & 163.35/4.50 & 280.00/2.70 \\ 
Po-Trip & 84.02/17.80 & 1.00/100.00 & 128.97/10.30 & 124.94/8.90 & 172.83/4.90 & 291.45/3.30 \\
RCE (Ours) & 16.95/45.50 & 1.01/98.80 & 22.67/39.80 & 41.85/29.50 & 71.47/14.30 & 165.36/8.90 \\ 

\bottomrule
\end{tabular}
}
\label{tab:targeted_transferability_ds121}
\end{table*}

\textbf{DenseNet121 as the surrogate model.}  In the main manuscript, we present the targeted transferability results with ResNet50 as the source model. Additionally, we choose DenseNet121 as the source white-box model. The results are shown in Table~\ref{tab:non-targeted_transferability_ds121} for non-targeted attack and in Table~\ref{tab:targeted_transferability_ds121} for targeted attack. 

\section{Top-$k$ Attack Strength is Transferable}

\begin{figure}[!htbp]
    \centering
    \includegraphics[width=0.9\linewidth]{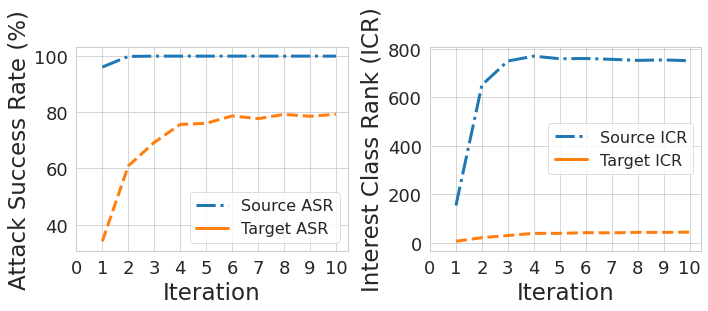}
    \caption{\textbf{Left:} Source ASR@1 and Target ASR@1 (average of several target networks) over iterations. \textbf{Right:} Source ICR and Target ICR (average of several target networks) over iterations.}
    \label{fig:source_target_asr_icr}
\end{figure}

\textbf{Transferability and Strength}
As was discussed in the main manuscript, source ASR@1 (source ICR) and target ASR@1 (target ICR) both increase over iterations. Now, we show this relationship more explicitly in Figure~\ref{fig:source_target_asr_icr}.

\begin{figure*}[!htbp]
    \centering
    \includegraphics[width=0.49\linewidth]{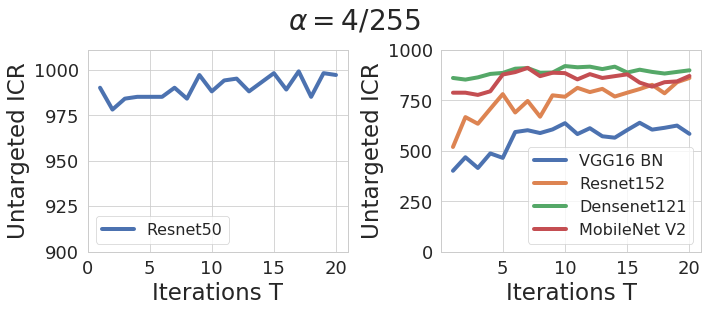}
    \includegraphics[width=0.49\linewidth]{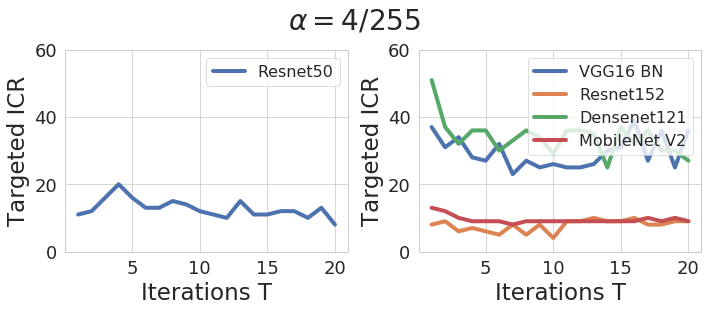}
    \caption{Untargeted (left two) and targeted (right two) ICR over $T$ iterations in both white-box (ResNet50) and black-box scenarios on a (randomly chosen) single sample.}
    \label{fig:untargeted_targeted_icr_single_sample}
\end{figure*}

\textbf{Single Sample Analysis.}
Figure~\ref{fig:untargeted_targeted_icr_single_sample} shows the results of ICR in untargeted and targeted settings over $T$ iterations for both white-box and black-box models for a (randomly chosen) single sample. The results show that top-$k$ attack strength based on the metric of ICR is transferable on a single sample (see the similar trend of ICR with more iterations on the white-box and black-box models).

\section{Zero Sum Constraint}
\begin{table}[!htbp]
\caption{Zero sum experiment results on the ImageNet dataset with various network architectures. The metrics reported are the average (with standard deviation) over the validation samples.}
\centering
\scalebox{0.67}{
\begin{tabular}{c|ccccccccc}
\toprule
& Sum & Abs. Sum & Std & Min & Max \\
\midrule
RN50 & 0.01±0.00 & 1830.44±284.83 & 2.45±0.36 & -5.80±0.93 & 16.87±4.70 \\
DN121 & 0.02±0.00 & 1876.99±291.23 & 2.50±0.36 & -6.23±1.01 & 15.96±4.14 \\
VGG16bn & 0.01±0.00 & 2324.19±410.37 & 3.09±0.56 & -6.76±1.26 & 19.12±6.55 \\
RN152 & 0.00±0.00 & 1825.42±302.96 & 2.45±0.38 & -5.84±0.97 & 17.67±4.58 \\
MNv2 & 0.08±0.06 & 2304.12±298.43 & 3.03±0.40 & -8.12±1.28 & 17.22±4.90 \\
\bottomrule
\end{tabular}
}
\label{tab:zero_sum_imagenet}
\end{table}

\begin{table}[!htbp]
\caption{Zero sum experiment results on the CIFAR10 (top) and CIFAR100 (bottom) datasets with various network architectures. The metrics reported are the average (with standard deviation) over the validation samples.}
\centering
\scalebox{0.6}{
\begin{tabular}{c|ccccccccc}
\toprule
& Sum & Abs. Sum & Std & Min & Max \\
\midrule
RN20 & 0.02±0.02 & 56.21±12.76 & 7.73±1.90 & -8.29±2.01 & 19.15±6.48 \\
RN32 & 0.02±0.01 & 45.75±8.84 & 6.48±1.27 & -6.46±1.57 & 16.60±4.60 \\
RN44 & 0.05±0.02 & 46.41±9.12 & 6.59±1.39 & -6.45±1.44 & 16.99±4.95 \\
RN56 & 0.03±0.02 & 39.99±7.88 & 5.78±1.10 & -5.49±1.43 & 15.15±3.86 \\
VGG19bn & 0.00±0.00 & 22.55±2.45 & 3.49±0.38 & -3.67±1.00 & 9.42±1.38 \\
DenseNet-BC-190-40 & 0.01±0.00 & 27.82±4.04 & 4.28±0.49 & -3.54±0.87 & 11.89±1.79 \\
\midrule
RN20 & 0.32±0.15 & 483.70±98.10 & 6.21±1.29 & -13.60±2.97 & 20.33±6.77 \\
RN32 & 0.20±0.13 & 511.02±93.67 & 6.58±1.24 & -14.07±2.87 & 22.71±7.19 \\
RN44 & 0.20±0.14 & 498.21±87.48 & 6.45±1.17 & -13.81±2.78 & 22.95±7.24 \\
RN56 & 0.29±0.17 & 474.85±79.69 & 6.15±1.08 & -13.00±2.60 & 22.39±6.95 \\
VGG19bn & 0.00±0.00 & 223.78±22.21 & 2.89±0.27 & -4.34±0.51 & 12.70±2.12 \\
DenseNet-BC-190-40 & 0.03±0.00 & 189.25±32.73 & 2.74±0.49 & -4.69±0.96 & 15.02±5.01 \\
\bottomrule
\end{tabular}
}
\label{tab:zero_sum_cifar10}
\end{table}

\begin{table}[t]
\caption{Zero sum experiment results for the adversarial images crafted with different losses: CW (top), CE (middle), RCE (bottom) on ImageNet dataset with ResNet50 (white-box model) and DenseNet121.} 
\centering
\scalebox{0.7}{
\begin{tabular}{c|ccccccccc}
\toprule
& Sum & Abs. Sum & Std & Min & Max \\
\midrule
RN50 & 0.01±0.00 & 1917.60±232.74 & 2.49±0.32 & -6.30±0.87 & 14.61±4.55 \\
DN121 & 0.02±0.00 & 2224.91±371.48 & 2.95±0.52 & -7.33±1.31 & 21.04±6.42 \\
\midrule
RN50 & 0.01±0.00 & 2001.94±251.94 & 2.63±0.36 & -6.49±0.93 & 16.91±5.61  \\
DN121 & 0.02±0.00 & 2373.04±419.24 & 3.19±0.61 & -7.70±1.51 & 24.05±7.69 \\
\midrule
RN50 & 0.01±0.00 & 1720.35±178.80 & 2.20±0.23 & -5.92±0.78 & 10.29±2.26 \\
DN121 & 0.02±0.00 & 1869.99±262.75 & 2.38±0.35 & -7.78±1.97 & 9.52±2.40 \\
\bottomrule
\end{tabular}
}
\label{tab:zero_sum_imagenet_adv_images}
\end{table}

\subsection{Phenomenon}
As introduced in the main manuscript, the \textit{zero-sum} phenomenon of logit vector $\mathbf{Z}$ shows that the sum of the logits mostly results in a value close to zero for both clean and adversarial samples. Here, we empirically demonstrate the phenomenon of the ``zero sum" constraint by evaluating the sum of the logit value $\mathbf{Z}$ on the ImageNet-compatible dataset introduced in the NeurIPS 2017 (See the Experimental general setup) and CIFAR10/CIFAR100 for different network architectures. 
From the results in Table~\ref{tab:zero_sum_imagenet} and Table~\ref{tab:zero_sum_cifar10}, the first observation is that the sum of all logit values in $\mathbf{Z}$, \ie $\sum_{i=1}^{i=K} z_{i}$ is indeed close to zero with a very small variance among the validation samples, indicating $\sum_{i=1}^{i=K} z_{i}$ is very close to zero for all validation samples. To further demonstrate that this phenomenon is occurring, not just due to very small values in $\mathbf{Z}$, we further present the absolute sum, \ie\ $\sum_{i=1}^{i=K} |z_{i}|$. Additionally, the relatively large values for the standard deviation, the minimum, and maximum value for the $Z$ statistics demonstrate that there exists a balance between the negative and positive logit values, which results in their sum being zero. This phenomenon is also observed for adversarial samples. We report the results for adversarial examples with different losses (CW, CE, RCE) on ImageNet dataset with ResNet50 transferring to DenseNet121 in Table~\ref{tab:zero_sum_imagenet_adv_images}. The results show that the zero-sum constraint also holds for adversarial examples.

\subsection{Influence of Network Weights Initialization}
We investigate the influence of the network weight initialization on the zero sum phenomenon. We observe that zero sum constraint is still present even with different weight initialization parameters suggesting that there is no influence of the network weight initialization on the phenomenon. The results are reported in Table~\ref{tab:zero_sum_cifar10_mean_0_std_1}, Table~\ref{tab:zero_sum_cifar10_mean_1_std_1}, Table~\ref{tab:zero_sum_cifar10_mean_3_std_3}, Table~\ref{tab:zero_sum_cifar10_mean_5_std_5}, and Table~\ref{tab:zero_sum_cifar100_mean_5_std_5}.

\begin{table}[t]
\caption{Zero sum experiment results on the CIFAR10 dataset with various network architectures. The metrics reported are the average (with standard deviation) over the validation samples. Weights of the network were initialized with $\sim \mathcal{N}(0,\,1^{2})$.}
\centering
\scalebox{0.6}{
\begin{tabular}{c|ccccccccc}
\toprule
Network & Sum & Abs. Sum & Std & Min & Max & Accuracy \\
\midrule
RN56 & 0.00±0.00 & 29.93±4.99 & 4.38±0.70 & -3.79±0.78 & 11.75±2.62 & 93.59\% \\
VGG19bn & 0.00±0.00	& 19.12±1.57 & 3.05±0.22 & -2.24±0.37 & 8.71±1.06 & 93.38\% \\
\bottomrule
\end{tabular}
}
\label{tab:zero_sum_cifar10_mean_0_std_1}
\end{table}

\begin{table}[!htbp]
\caption{Zero sum experiment results on the CIFAR10 dataset with various network architectures. The metrics reported are the average (with standard deviation) over the validation samples. Weights of the network were initialized with $\sim \mathcal{N}(1,\,1^{2})$.}
\centering
\scalebox{0.6}{
\begin{tabular}{c|ccccccccc}
\toprule
Network & Sum & Abs. Sum & Std & Min & Max & Accuracy \\
\midrule
RN56 & 0.00±0.00 & 32.13±5.81 & 4.59±0.78 & -4.24±0.92 & 12.01±2.81 & 92.48\% \\
VGG19bn & 0.00±0.00 & 22.86±2.73 & 3.44±0.40 & -3.63±0.55 & 9.19±1.41 & 93.04\% \\
\bottomrule
\end{tabular}
}
\label{tab:zero_sum_cifar10_mean_1_std_1}
\end{table}

\begin{table}[!htbp]
\caption{Zero sum experiment results on the CIFAR10 dataset with various network architectures. The metrics reported are the average (with standard deviation) over the validation samples. Weights of the network were initialized with $\sim \mathcal{N}(3,\,3^{2})$.}
\centering
\scalebox{0.6}{
\begin{tabular}{c|ccccccccc}
\toprule
Network & Sum & Abs. Sum & Std & Min & Max & Accuracy \\
\midrule
RN56 & 0.00±0.00 & 39.15±8.08 & 5.44±1.21 & -6.86±2.69 & 13.09±3.76 & 91.77\% \\
VGG19bn & 0.00±0.00 & 25.40±3.94 & 3.87±0.54 & -6.35±2.01 & 9.20±1.28 & 92.70\% \\
\bottomrule
\end{tabular}
}
\label{tab:zero_sum_cifar10_mean_3_std_3}
\end{table}

\begin{table}[!htbp]
\caption{Zero sum experiment results on the CIFAR10 dataset with various network architectures. The metrics reported are the average (with standard deviation) over the validation samples. Weights of the network were initialized with $\sim \mathcal{N}(5,\,5^{2})$.}
\centering
\scalebox{0.6}{
\begin{tabular}{c|ccccccccc}
\toprule
Network & Sum & Abs. Sum & Std & Min & Max & Accuracy \\
\midrule
RN56 & 0.00±0.00 & 31.46±6.77 & 4.30±0.93 & -4.50±1.09 & 10.58±3.20 & 90.20\% \\
VGG19bn & 0.00±0.00 & 20.87±2.44 & 3.24±0.35 & -3.28±0.55 & 8.81±1.28 & 92.62\% \\
\bottomrule
\end{tabular}
}
\label{tab:zero_sum_cifar10_mean_5_std_5}
\end{table}

\begin{table}[!htbp]
\caption{Zero sum experiment results on the CIFAR100 dataset with various network architectures. The metrics reported are the average (with standard deviation) over the validation samples. Weights of the network were initialized with $\sim \mathcal{N}(5,\,5^{2})$.}
\centering
\scalebox{0.6}{
\begin{tabular}{c|ccccccccc}
\toprule
Network & Sum & Abs. Sum & Std & Min & Max & Accuracy \\
\midrule
RN56 & 0.01±0.00 & 377.77±75.17 & 4.79±0.96 & -10.98±2.36 & 14.10±4.35 & 63.83\% \\
VGG19bn & 0.00±0.00 & 558.74±83.81 & 6.78±0.98 & -15.31±2.52 & 17.14±2.80 & 68.32\% \\
\bottomrule
\end{tabular}
}
\label{tab:zero_sum_cifar100_mean_5_std_5}
\end{table}

\subsection{Influence of Unbalanced Dataset}
Additionally, we confirm that the zero sum constraint is valid when the dataset classes are unbalanced. For 10 classes in CIFAR10, we test two variants of unbalance. In the first setup, we set the number of samples for each class to change linearly (from 5000 to 500). The result is in Table~\ref{tab:zero_sum_cifar10_unbalanced_linear}. In the second setup, we adopt the unbalanced class setting as the common long-tail problem setup~\cite{cui2019class} with the imbalance factor set to 50 or 100. The results for this setup are in Table~\ref{tab:zero_sum_cifar10_imbalance_factor_50} and Table~\ref{tab:zero_sum_cifar10_imbalance_factor_100}. The result for CIFAR100 dataset with imbalance factor of 50 is in Table~\ref{tab:zero_sum_cifar100_imbalance_factor_50}.

\begin{table}[t]
\caption{Zero sum experiment results on the unbalanced CIFAR10 dataset (linear change) with various network architectures. The metrics reported are the average (with standard deviation) over the validation samples.}
\centering
\scalebox{0.6}{
\begin{tabular}{c|ccccccccc}
\toprule
Network & Sum & Abs. Sum & Std & Min & Max & Accuracy \\
\midrule
RN56 & 0.00±0.00 & 25.70±4.81 & 3.82±0.73 & -2.99±0.73 & 10.41±2.60 & 91.22\% \\
VGG19bn & 0.00±0.00 & 20.55±2.31 & 3.24±0.37 & -3.60±0.73 & 8.70±1.35 & 89.89\% \\
\bottomrule
\end{tabular}
}
\label{tab:zero_sum_cifar10_unbalanced_linear}
\end{table}

\begin{table}[!htbp]
\caption{Zero sum experiment results on the unbalanced CIFAR10 dataset (imbalance factor 50) with various network architectures. The metrics reported are the average (with standard deviation) over the validation samples.}
\centering
\scalebox{0.6}{
\begin{tabular}{c|ccccccccc}
\toprule
Network & Sum & Abs. Sum & Std & Min & Max & Accuracy \\
\midrule
RN56 & 0.00±0.00 & 26.37±5.27 & 3.77±0.92 & -3.21±0.80 & 9.84±3.32 & 77.70\% \\
VGG19bn & 0.00±0.00 & 23.38±5.01 & 3.39±0.59 & -3.38±1.07 & 8.79±1.85 & 78.69\% \\
\bottomrule
\end{tabular}
}
\label{tab:zero_sum_cifar10_imbalance_factor_50}
\end{table}

\begin{table}[!htbp]
\caption{Zero sum experiment results on the unbalanced CIFAR10 dataset (imbalance factor 100) with various network architectures. The metrics reported are the average (with standard deviation) over the validation samples.}
\centering
\scalebox{0.6}{
\begin{tabular}{c|ccccccccc}
\toprule
Network & Sum & Abs. Sum & Std & Min & Max & Accuracy \\
\midrule
RN56 & 0.01±0.01 & 25.92±4.79 & 3.68±0.84 & -3.10±0.73 & 9.57±3.12 & 70.61\% \\
VGG19bn & 0.00±0.00 & 23.70±5.30 & 3.44±0.69 & -3.85±1.53 & 8.63±2.04 & 70.59\% \\
\bottomrule
\end{tabular}
}
\label{tab:zero_sum_cifar10_imbalance_factor_100}
\end{table}

\begin{table}[!htbp]
\caption{Zero sum experiment results on the unbalanced CIFAR100 dataset (imbalance factor 50) with various network architectures. The metrics reported are the average (with standard deviation) over the validation samples.}
\centering
\scalebox{0.6}{
\begin{tabular}{c|ccccccccc}
\toprule
Network & Sum & Abs. Sum & Std & Min & Max & Accuracy \\
\midrule
RN56 & 0.07±0.04 & 246.12±49.72 & 3.21±0.65 & -6.49±1.48 & 11.88±3.92 & 44.46\% \\
VGG19bn & 0.01±0.01 & 225.33±32.34 & 2.93±0.39 & -4.26±0.64 & 12.08±2.40 & 45.40\% \\
\bottomrule
\end{tabular}
}
\label{tab:zero_sum_cifar100_imbalance_factor_50}
\end{table}

\begin{figure}[!htbp]
    \centering
    \includegraphics[width=0.9\linewidth]{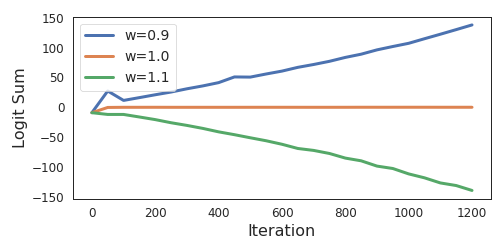}
    \caption{Influence of $w$ on the $\sum_{i=1}^{i=K} z_{i}$ in the training stage.}
    \label{fig:analysis_weight}
\end{figure}

\subsection{Possible Explanation}
Admittedly, we do not have a clear explanation for this phenomenon. Here, we only attempt to provide a possible explanation. Note that the DNN is often trained with the CE loss. Taking a closer look at the derivation of the CE loss with respect to the logit vector, \ie $\frac{\partial L_{CE}}{\partial \mathbf{Z}} = \mathbf{P} - \mathbf{Y}_{gt}$, it can be observed that sum of all values in both $\mathbf{P}$ and ${Y}_{gt}$ is 1. 
We believe that this constitutes a \textit{necessary} condition for making the $\sum_{i=1}^{i=K} z_{i}$ close to zero. To verify this claim, we experiment with a new loss that results in $\frac{\partial L}{\partial \mathbf{Z}} = w\mathbf{P} - \mathbf{Y}_{gt}$. When $w$ is set to a value larger than $1$, such as $1.1$, the loss makes the $\sum_{i=1}^{i=K} z_{i}$ smaller and smaller as the network training goes on (See Figure~\ref{fig:analysis_weight}). Similarly, when $w$ is set to a value smaller than $1$, such as $0.9$, the loss tends to increase the $\sum_{i=1}^{i=K} z_{i}$. In the above two cases, we observe that $\sum_{i=1}^{i=K} z_{i}$ eventually becomes infinitely negative/positive given enough iterations and consequently the network training does not converge. When $w$ is set to 1, which is identical to the original CE loss, we observe that $\sum_{i=1}^{i=K} z_{i}$ converges to zero. Overall, we find that   the sum of all values in $\frac{\partial L_{CE}}{\partial \mathbf{Z}}$ is a \textit{necessary}, but probably not \textit{sufficient}, condition for making $\sum_{i=1}^{i=K} z_{i}$ approach zero. We leave a more elaborate explanation for this phenomenon to future work. 

\section{Logit Vector Gradient Derivations}
Here, we provide a detailed derivation of the partial derivative of various loss functions with respect to the logit vector $Z$, shown in the main manuscript.  
\subsection{CE Loss}

Before demonstrating the derivative for the CE loss, we will first calculate the derivatives of the softmax output ($\mathbf{P}$) with respect to its input (the logit vector $\mathbf{Z}$). Each entry of the logit vector $\mathbf{Z}$ is indicated with index $i$, while each entry of the $\mathbf{P}$ is indicated with index $j$. For simplicity, we divide into two scenarios, $j = i$ and $j \neq i$, and conduct the derivation respectively. First, let's consider \ie $j=i$, and the derivative $\frac{\partial p_{j}}{\partial z_{i}}$ can be calculated as follows:
\begin{equation}
\label{eq5}
\begin{split}
\frac{\partial p_{i}}{\partial z_{i}} &= \frac{\partial (\frac{e^{z_i}}{\sum^{K}_{k=1}e^{z_{k}}})}{\partial {z}_{i}}\\ 
&= \frac{e^{z_i}\sum^{K}_{k=1}e^{z_{k}} - e^{2z_i}}{(\sum^{K}_{k=1}e^{z_{k}})^2}\\
&= p_{i} - p_{i}^2\\
&= p_{i}(1-p_{i}),
\end{split}
\end{equation}
with $p_{i}=\frac{e^{z_i}}{\sum_{k=1}^K e^{z_k}}$.
Eq.~\ref{eq5} enables us to obtain the derivative of the CE loss, \ie\ $L_{CE} = -\log p_{gt}$, with respect to the softmax input which has the ground-truth index, \ie $i=j=gt$:
\begin{equation}
\label{eq8}
\begin{split}
\frac{\partial L_{CE}}{\partial z_{gt}} &= \frac{\partial (-\log p_{gt})}{\partial {z}_{gt}} \\ 
&= -\frac{1}{p_{gt}}\frac{\partial p_{gt}}{\partial z_{gt}} \\
&= -\frac{1}{p_{gt}} (p_{gt}(1-p_{gt}) \\
&= p_{gt}-1.
\end{split}
\end{equation}

On the other hand, for the case when $j \neq i$, the derivative $\frac{\partial p_{j}}{\partial z_{i}}$ can be calculated as follows:
\begin{equation}
\label{eq6}
\begin{split}
\frac{\partial p_{j}}{\partial z_{i}} &= \frac{\partial (\frac{e^{z_j}}{\sum^{K}_{k=1}e^{z_{k}}})}{\partial {z}_{i}} \\ 
&= -\frac{e^{z_j}e^{z_i}}{(\sum^{K}_{k=1}e^{z_{k}})^2}\\
&= -p_{j}p_{i}.
\end{split}
\end{equation}
With Eq.~(\ref{eq6}), we further calculate the derivative of the CE loss with respect to the softmax inputs which are different from the ground-truth index, \ie $i \neq gt$: 
\begin{equation}
\label{eq7}
\begin{split}
\frac{\partial L_{CE}}{\partial z_{i}} &= \frac{\partial (-\log p_{gt})}{\partial {z}_{i}} \\ 
&= -\frac{1}{p_{gt}}\frac{\partial p_{gt}}{\partial z_i} \\
&= -\frac{1}{p_{gt}} (-p_{gt}p_{i}) \\
&= p_{i}.
\end{split}
\end{equation}

From Eq.~\ref{eq8} and Eq.~\ref{eq7}, we arrive at the formulation presented in the main manuscript:
\begin{equation}
\label{eq9}
\begin{split}
\frac{\partial L_{CE}}{\partial \mathbf{Z}} = \mathbf{P} - \mathbf{Y}_{gt},
\end{split}
\end{equation}
with $\mathbf{Y}_{i}$ indicating a one-hot encoded vector with the position at index $i$ being one. Thus, the derivative of the CE(LL) loss, \ie\ $L_{CE} = \log P_{LL}$,  to the logit vector can be derived similarly with the final formulation as:
\begin{equation}
\label{eq10}
\begin{split}
\frac{\partial L_{CE(LL)}}{\partial \mathbf{Z}} = \mathbf{Y}_{LL} - \mathbf{P}.
\end{split}
\end{equation}

\subsection{CW Loss}
CE and CW loss are the two most widely used losses for the white-box attack~\cite{gowal2019alternative,lee2020adversarial}. In the above, we derive the gradient for CE loss and we further conduct a similar derivation for CW loss which is denoted as $L_{CW} = z_{j} - z_{gt}$ ~\cite{gowal2019alternative,lee2020adversarial} with $j = \arg \max\limits_{i \neq gt} z_{i} $ indicating the highest class except for the gt class. The derivative of the $L_{CW}$ to the $Z$ is denoted as $\frac{\partial L_{CW}}{\partial \mathbf{Z}}$. $\frac{\partial L_{CW}}{\partial z_i} = 0$  
when $i \neq j$ and $i \neq gt$. $\frac{\partial L_{CW}}{\partial z_i}$  is $1$ and $-1$ when  $i = j$ and $i = gt$, respectively.  Therefore, we arrive at:
\begin{equation}
\label{eq:cw}
\begin{split}
\frac{\partial L_{CW}}{\partial \mathbf{Z}} = \mathbf{Y}_{j} - \mathbf{Y}_{gt}.
\end{split}
\end{equation}

\subsection{Relative Cross-Entropy (RCE) Loss}
With Eq.~\ref{eq9}, we can calculate the derivative of the proposed RCE loss:
\begin{equation}
\begin{split}
\label{eq12}
\frac{\partial L_{RCE}}{\partial \mathbf{Z}} &= \frac{\partial (L_{CE_{gt}}-\frac{1}{K}\sum^{K}_{k=1}L_{CE_{k}})}{\partial \mathbf{Z}} \\
&= \frac{\partial L_{CE_{gt}}}{\partial \mathbf{Z}} - \frac{1}{K}\sum^{K}_{k=1}\frac{\partial L_{CE_{k}}}{\partial \mathbf{Z}} \\
&= \mathbf{P} - \mathbf{Y_{gt}} - \frac{1}{K}\sum^{K}_{k=1}(\mathbf{P} - \mathbf{Y_{k}}) \\
&= \frac{1}{K}\mathbf{1} - \mathbf{Y}_{gt}.
\end{split}
\end{equation}
where $\mathbf{1}$ indicates a vector with all values being $1$.

\section{CW and RCE are Special Cases of CE}
\subsection{Derivative of the Temperature Scaled CE-loss}
The derivative of the CE-Loss with temperature scaling can be written as:

\begin{equation}
\label{CE_withT}
\frac{\partial L_{CE(Temp)}}{\partial \textbf{Z}} = \frac{1}{T_e} (\mathbf{P}_e - \mathbf{Y}_{gt}), 
\end{equation}
This derivation unfolds similarly to the one previously presented for the CE Loss without temperature. Each entry of the logit vector $\mathbf{Z}$ is indicated with index $i$, while each entry of $\mathbf{P}$ is indicated with index $j$. Again first looking at the softmax output with $T_e$ ($\mathbf{P}_e$) with respect to the logit vector $\mathbf{Z}$ with $i=j$ we arrive at:
\begin{equation}
\begin{split}
\frac{\partial p_{e}^i}{\partial z_{i}} &= \frac{\partial (\frac{e^{z_i/T_e}}{\sum^{K}_{k=1}e^{z_{k}/T_e}})}{\partial {z}_{i}}\\ 
&= \frac{1}{T_e}(p_{e}^i(1-p_{e}^i)).
\end{split}
\end{equation}
For the case where $i \neq j$ we arrive at the following derivative:

\begin{equation}
\begin{split}
\frac{\partial p_{e}^j}{\partial z_{i}} &= \frac{\partial (\frac{e^{z_j/T_e}}{\sum^{K}_{k=1}e^{z_{k}}})}{\partial {z}_{i}} \\ 
&= \frac{1}{T_e}(-p_{e}^{j} p_{e}^{i}).
\end{split}
\end{equation}
Analogous to Eq.~(\ref{eq8}) and Eq.~(\ref{eq7}), we can calculate the derivatives for the CE Loss with $T_e$.
For the case $i= gt$ we arrive at:
\begin{equation}
\begin{split}
\frac{\partial L_{CE(Temp)}}{\partial z_{gt}} &= \frac{\partial (-\log p_{e}^{gt})}{\partial {z}_{gt}} \\ 
&= \frac{1}{T_e} (p_{e}^{gt}-1).
\end{split}
\label{eq14}
\end{equation}
For the case $i \neq gt$ we arrive at:
\begin{equation}
\begin{split}
\frac{\partial L_{CE(Temp)}}{\partial z_{i}} &= \frac{\partial (-\log p_{e}^{gt})}{\partial {z}_{i}} \\ 
&= \frac{1}{T_e} p_{e}^{i},
\end{split}
\label{eq13}
\end{equation}

With Eq.~(\ref{eq14}) and  Eq.~(\ref{eq13}), we finally arrive at Eq.~(\ref{CE_withT}).

\subsection{Scale-invariant Property of the Gradient Derivative}
As highlighted in the main manuscript, only the direction of the derivative matters and the scale is irrelevant because FGSM is adopted as the basic method for all approaches to get the sign of the derivative. Without losing generality, we compare two losses $L_{A}$ and $L_{B}$ by setting $\frac{\partial  L_{B}}{\partial \mathbf{Z}} = s \frac{\partial L_{A}}{\partial \mathbf{Z}}$ where $s$ is a scale factor.  We can derive:
\begin{equation}
\label{eq:scale_invariant}
\begin{split}
sign(\frac{\partial L_{B}}{\partial \mathbf{X}}) &= sign(\frac{\partial \mathbf{Z}}{\partial \mathbf{X}} \frac{\partial L_{B}}{\partial \mathbf{Z}})\\ 
&= sign(s \frac{\partial \mathbf{Z}}{\partial \mathbf{X}}  \frac{\partial L_{A}}{\partial \mathbf{Z}})\\ 
&= sign(\frac{\partial \mathbf{Z}}{\partial \mathbf{X}}  \frac{\partial L_{A}}{\partial \mathbf{Z}})\\ 
&= sign(\frac{\partial L_{A}}{\partial \mathbf{X}}) 
\end{split}
\end{equation}

\subsection{Relationship to Other Loss Functions}
The probability of the $i$-th class in $\mathbf{P}_e$ is shown as:
\begin{equation}
\label{eq2}
p_{e}^{i} = \frac{e^{z_{i}/T_{e}}}{\sum_{k=1}^{k=K} e^{z_{k}/T_{e}}} 
\end{equation}
Note that $T_{e}$ ranges from $(0, \infty)$. Without losing generality, by assuming $z_{x} > z_{y}$, we can derive:
\begin{equation}
\begin{split}
\label{eq3}
\frac{p_{e}^{x}}{p_{e}^{y}} &= \frac{\frac{e^{z_{x}/T_{e}}}{\sum_{k=1}^{k=K} e^{z_{k}/T_{e}}}}{\frac{e^{z_{y}/T_{e}}}{\sum_{k=1}^{k=K} e^{z_{k}/T_{e}}}} \\
&= \frac{e^{z_{x}/T_{e}}}{e^{z_{y}/T_{e}}} \\
&= e^{(z_{x} - z_{y})/T_{e}} \\
&> 1
\end{split}
\end{equation}
\textbf{RCE Loss can be seen as a Special Case of CE Loss.} For $z_{x} > z_{y}$ and $T_e \rightarrow \infty$, we can derive:

\begin{equation}
\begin{split}
\label{eq:3}
\lim_{T_e \rightarrow \infty}\frac{p_{e}^{x}}{p_{e}^{y}} &= \lim_{T_e \rightarrow \infty}e^{(z_{x} - z_{y})/T_{e}} \\
&= 1
\end{split}
\end{equation}
With the above equation and $\sum_{k=1}^{k=K} p_{e}^{i} = 1$, it can be concluded that $\mathbf{P}_e = \frac{1}{K} \mathbf{1}$ when $T_e \rightarrow \infty$ or when $T_e$ is set to a large value.
Thus, in this case, Eq.~(\ref{CE_withT}) can be further derived as follows:
\begin{equation}
\begin{split}
\label{eq:rce_equivalent}
\frac{\partial L_{CE(Temp)}}{\partial \textbf{Z}} &= \frac{1}{T_e} (\mathbf{P}_e - \mathbf{Y}_{gt}) \\
&= \frac{1}{T_e} (\frac{1}{K} \mathbf{1} - \mathbf{Y}_{gt})
\end{split}
\end{equation}
Given the scale-invariant property indicated by Eq.~(\ref{eq:scale_invariant}), Eq.~(\ref{eq:rce_equivalent}) is equivalent to the derived gradient in Eq.~(\ref{eq12}) for the RCE loss. Thus, we conclude that the RCE loss can be seen as a special case of the CE loss by setting $T_{e}$ to a large value.

\textbf{CW loss can be seen as a Special Case of CE Loss.} We will now show the behavior of $\mathbf{P_{e}^{i}}$ when $T_e \rightarrow 0$. Given $z_{x} > z_{y}$, we can derive:
\begin{equation}
\begin{split}
\label{eq:approach_zero}
\lim_{T_e \rightarrow 0}\frac{P_{e}^{x}}{P_{e}^{y}} &= \lim_{T_e \rightarrow 0}e^{(z_{x} - z_{y})/T_{e}} \\
&= \infty
\end{split}
\end{equation}

If $i_{max}$ is the index of the class with the largest logit, $\lim_{T_e \rightarrow 0}p_{e}^{i_{max}} = 1$. Otherwise, $\lim_{T_e \rightarrow 0} p_{e}^{i} = 0$ ($i \neq i_{max}$).  Given the definition $j = \arg \max\limits_{i \neq gt} z_{i}$, we know that the class with the highest logit in $\textbf{Z}$ is either the j-th class or the gt class. Thus, for small enough $T_e$ ($T_e \rightarrow 0$), $p_{e}^{gt} + p_{e}^{j} = 1$.
Let us denote $p_{j} = m$ and $p_{gt} = 1-m$. 
Then, Eq.~\ref{CE_withT} can be rewritten as 
\begin{equation}
\label{eq:cw_equivalent}
\frac{\partial L_{CE(Temp)}}{\partial \textbf{Z}} = \frac{m}{T_{e}}(\mathbf{Y}_{j} - \mathbf{Y}_{gt})  , 
\end{equation}
Given the scale-invariant property indicated by Eq.~\ref{eq:scale_invariant}, Eq.~\ref{eq:cw_equivalent} is equivalent to the derived gradient in Eq.~\ref{eq:cw} for the CW loss. Thus, we conclude that the CW loss can be seen as a special case of the CE loss by setting $T_{e}$ to a very small value.

\begin{figure}[!htbp]
    \centering
    \scalebox{0.95}{
    \includegraphics[width=\linewidth]{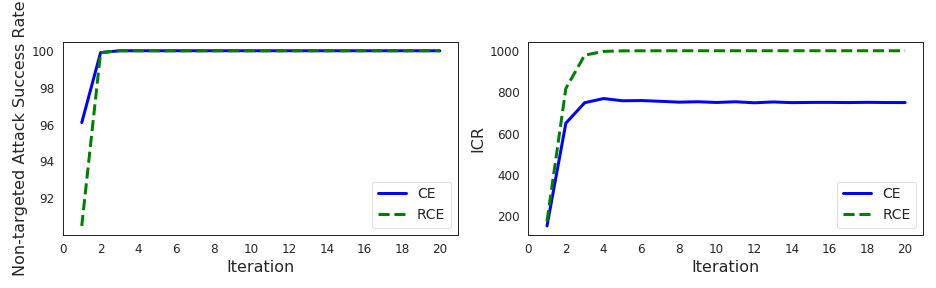}
    }
    \caption{Non-targeted attack success rate and ICR on the white-box ResNet50.}
    \label{fig:icr_non_targeted_acc}
\end{figure}

\begin{figure}[!htbp]
    \centering
    \scalebox{0.95}{
    \includegraphics[width=\linewidth]{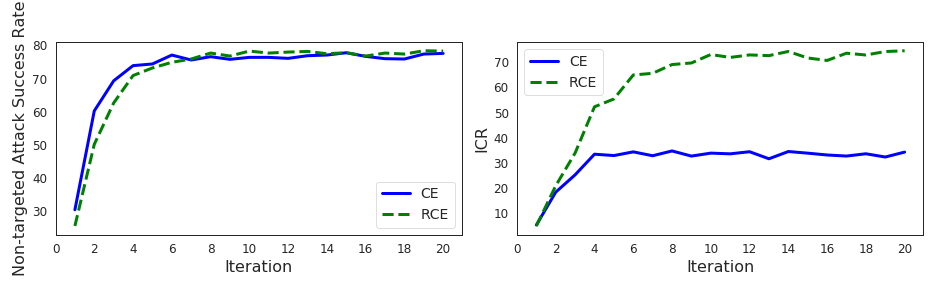}
    }
    \caption{Non-targeted attack success rate and ICR on the black-box DenseNet121.}
    \label{fig:icr_non_targeted_acc_transfer}
\end{figure}

\section{Limitation of RCE Loss in the Early Iterations}

As indicated in the main manuscript, the proposed RCE loss might converge slower than the existing CE loss due to its position-agnostic property. Transferring from ResNet50 to DenseNet121 on the ImageNet, we provide the white-box results and black-box results in Figure~\ref{fig:icr_non_targeted_acc} and Figure~\ref{fig:icr_non_targeted_acc_transfer}, respectively. We observe that in the early iterations, CE outperforms our proposed RCE loss, especially for the metric of attack success rate.

\end{document}